\documentclass[11pt]{article}

\usepackage[final]{acl}

\usepackage{times}
\usepackage{latexsym}

\usepackage[T1]{fontenc}

\usepackage[utf8]{inputenc}

\usepackage{microtype}

\usepackage{inconsolata}

\usepackage{graphicx}
\usepackage{booktabs}
\usepackage{multirow}
\usepackage{subcaption}
\usepackage{tcolorbox}
\usepackage{siunitx}
\usepackage[table]{xcolor}

\newtcolorbox{promptbox}{
  colback=gray!5,      
  colframe=gray!50,    
  arc=2mm,             
  boxrule=0.3pt,       
  left=4pt, right=4pt, top=4pt, bottom=4pt, 
  before skip=6pt, after skip=6pt,          
}
\usepackage{enumitem}
\usepackage{listings}
\title{Test-Time Scaling of Reasoning Models for Machine Translation}

\author{Zihao Li,$^{1}$~Shaoxiong Ji,$^{2,3,1}$~Jörg Tiedemann$^{1}$ \\
$^{1}$University of Helsinki \quad  $^{2}$ ELLIS Institute Finland \quad  $^{3}$ University of Turku
\\
\texttt{firstname.lastname@\{$^{1}$helsinki.fi,~$^{3}$utu.fi\}}
}

\begin{document}

\maketitle
\begin{abstract}
Test-time scaling (TTS) has enhanced the performance of Reasoning Models (RMs) on various tasks such as math and coding, yet its efficacy in machine translation (MT) remains underexplored. 
This paper investigates whether increased inference-time computation improves translation quality. 
We evaluate 12 RMs across a diverse suite of MT benchmarks spanning multiple domains, examining three scenarios: direct translation, forced-reasoning extrapolation, and post-editing. 
Our findings show that for general-purpose RMs, TTS provides limited and inconsistent benefits for direct translation, with performance quickly plateauing. 
However, the effectiveness of TTS is unlocked by domain-specific fine-tuning, which aligns a model's reasoning process with task requirements, leading to consistent improvements up to an optimal, self-determined reasoning depth. 
We also find that forcing a model to reason beyond its natural stopping point consistently degrades translation quality. 
In contrast, TTS proves highly effective in a post-editing context, reliably turning self-correction into a beneficial process. 
These results indicate that the value of inference-time computation in MT lies not in enhancing single-pass translation with general models, but in targeted applications like multi-step, self-correction workflows and in conjunction with task-specialized models.
\end{abstract}

\section{Introduction}
\begin{figure}[ht]
    \centering
    \includegraphics[width=\columnwidth]{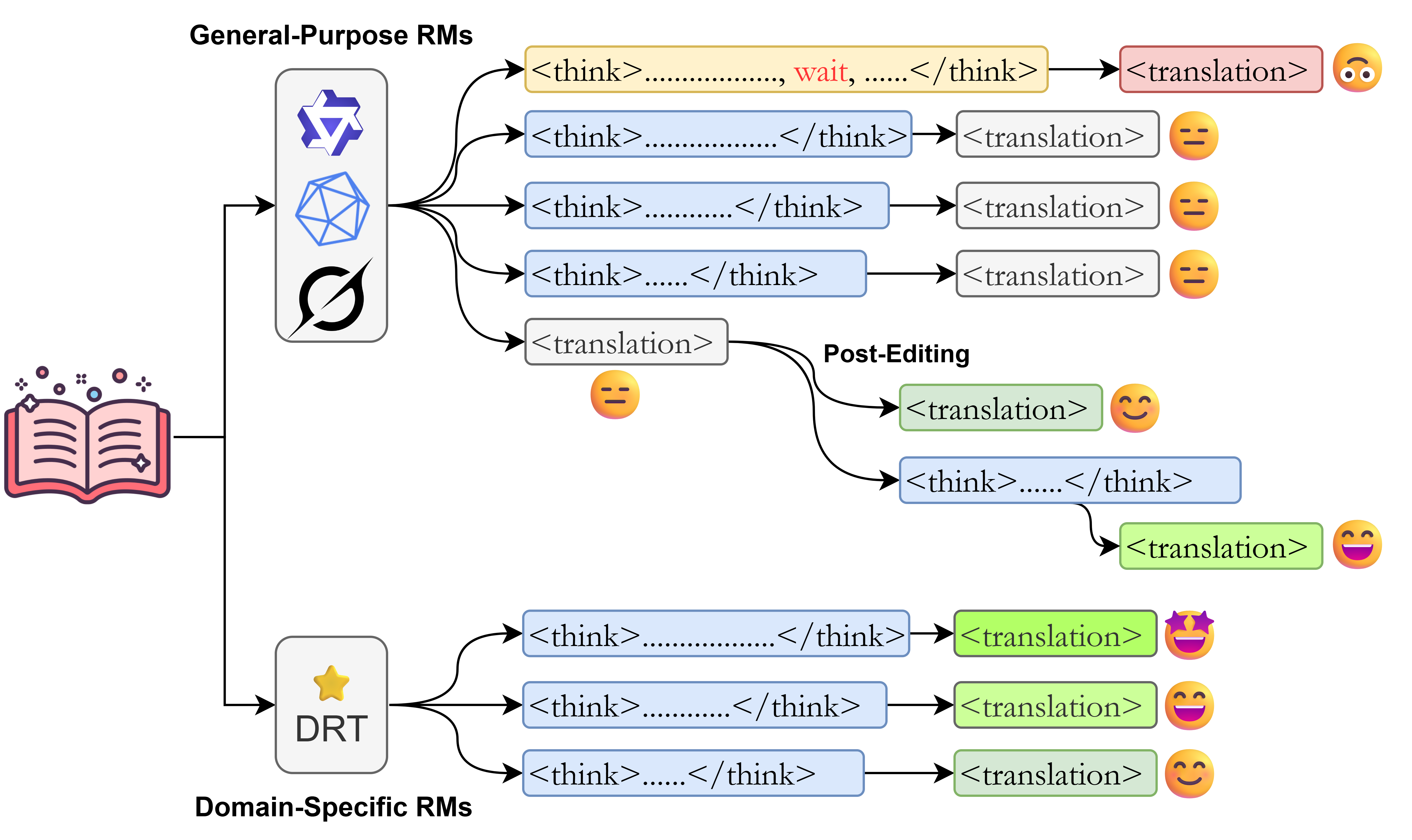}
    \caption{Illustration of the effectiveness of test-time scaling in reasoning models for machine translation. (1) TTS for general-purpose RMs yields only a small initial performance gain, but quickly plateauing as increased inference cost. (2) Forcing RMs to reason beyond their natural stopping point degrades quality by introducing noise. (3) In contrast, TTS becomes effective when applied to RMs specifically developed for MT. (4) TTS shows improvements in post-editing workflows. All these highlight TTS's value in MT lies in task-specialized models and multi-step self-correction, rather than as a robust strategy for enhancing single-pass translation with general-purpose RMs.}
    \label{fig:rm4mt}
\end{figure}
Large language models (LLMs) have dramatically advanced machine translation (MT), evolving from statistical and neural paradigms to systems capable of handling diverse languages, domains, and complexities with unprecedented accuracy~\citep{lyu2023paradigm, kocmi-etal-2024-findings, zhu-etal-2024-multilingual, cui-etal-2025-multilingual, hendy2023good}.
Recent developments in Reasoning Models (RMs), models designed to incorporate structured reasoning processes like Chain-of-Thought (CoT), have further transformed MT by reframing it as a cognitive task requiring contextual analysis, cultural adaptation, and self-reflection~\citep{liu2025new}. 
For instance, RMs can resolve ambiguities in stylized texts and maintain coherence across documents, thereby outperforming traditional LLMs in semantically demanding scenarios~\citep{ye2025well}.

Test-time scaling (TTS) has emerged as a transformative approach for enhancing model performance, which allocates additional computational resources during inference to enhance performance without requiring model retraining or parameter expansion~\citep{snell2024scaling}.
The effectiveness of TTS has been particularly pronounced for RMs such as DeepSeek-R1~\citep{Guo_2025}, Gemini 2.5~\citep{comanici2025gemini}, and OpenAI's o-Series~\citep{jaech2024openai, openai_o3_o4mini_system_card_2025}, which have achieved breakthrough performance on challenging benchmarks by extending their reasoning chains. 
Moreover, relatively small RMs have demonstrated impressive results on mathematical and coding tasks through TTS~\citep{muennighoff2025s1, li2025s}, suggesting that inference-time computation can partially compensate for limited model capacity.

Nevertheless, applying TTS to RMs for MT introduces distinct challenges and untapped potential that warrant deeper exploration.
Unlike math or coding tasks, where correctness can often be objectively determined, MT demands not only linguistic accuracy but also reasoning over cultural nuances, domain-specific terminology, and long-range dependencies, areas where unstructured compute scaling may yield diminishing returns. 
Moreover, interventions like forced extrapolation (e.g., inserting ``\texttt{wait}" tokens to extend reasoning) could disrupt natural deliberation, potentially introducing noise. 
In post-editing (PE) contexts, where models refine their own drafts, TTS might unlock iterative improvements, though this demands rigorous testing across varied benchmarks.

This paper investigates these open questions by distinguishing between two TTS workflows: \textit{Direct Translation} (single-pass CoT scaling) and \textit{Post-Editing} (compute-scaled self-correction). We structure our investigation through three research questions:
\begin{itemize}[nolistsep,noitemsep]
    \item \textbf{RQ1: How effective is test-time scaling for MT?} We examine whether increased inference computation reliably boosts translation quality across general-purpose and fine-tuned MT-specific RMs.
    \item \textbf{RQ2: Does extrapolation by inserting ``\texttt{wait}'' forcibly help?} We investigate if overriding models' natural stopping points, which further scales up the inference computation, enhances or hinders performance.
    \item \textbf{RQ3: Does test-time scaling work in post-editing?} We evaluate TTS in self-correction scenarios, assessing its role in refining initial translations when being allocated with specific compute budgets.
\end{itemize}

To address these questions, we assemble a comprehensive array of MT benchmarks encompassing literary, biomedical, cultural, commonsense, constrained terminology, and retrieval-augmented domains. We assess 12 RMs, spanning open-source series (Qwen-3~\citep{yang2025qwen3technicalreport}, Cogito~\citep{deepcogito_cogito_v1_preview}, DRT~\citep{wang-etal-2025-drt}) and the proprietary Grok-3-Mini. 

Our key contributions and findings are as follows:
\begin{itemize}[nolistsep,noitemsep]
    \item We demonstrate that for general-purpose RM, TTS provides limited and inconsistent benefits for direct machine translation.
    After small initial improvements at very low budgets, performance plateaus across metrics and datasets, indicating that ``more thinking'' alone is not a robust path to better translations.
    \item We show that the effectiveness of TTS is unlocked by domain-specific fine-tuning, which aligns the model’s reasoning process with task requirements.
    For DRT models fine-tuned on specific domain data, performance improves with budget on in-domain tasks and saturates once models naturally stop increasing their internal token usage, suggesting an emergent alignment between optimal reasoning depth and task demands. 
    This alignment largely disappears out of the domain.
    \item We find that forcing a model to reason by inserting a single ``\texttt{wait}'' beyond its natural stopping point consistently degrades translation quality, highlighting the importance of the model's intrinsic deliberation process.
    \item We establish that TTS is highly effective in a post-editing context, in which the inference cost is higher than the cost of direct translation. TTS turns self-correction into a reliably beneficial process.
\end{itemize}
These findings provide implications for deploying RMs in production MT systems and highlight the critical interplay between model capacity, task-specific training, and inference-time computation in determining when and how test-time scaling benefits translation quality.

\section{Experimental Setup}

\subsection{Datasets}
To comprehensively evaluate the reasoning capabilities and scaling properties of models at test time, we curated a diverse suite of eight machine translation benchmarks. 
These datasets span multiple domains, granularities, and languages, targeting a wide spectrum of reasoning challenges (Table~\ref{tab:datasets}).

For tasks requiring deep contextual and stylistic understanding, we use three literary benchmarks: the document-level \textbf{WMT24-Literary}~\citep{wang-etal-2024-findings}, paragraph-level \textbf{LitEval-Corpus}~\citep{zhang-etal-2025-good}, and sentence-level \textbf{MetaphorTrans}~\citep{wang-etal-2025-drt}. 
These datasets are rich in complex linguistic phenomena, cultural references, and figurative language (similes and metaphors), demanding sophisticated reasoning to preserve literary style and meaning.
Similarly, the \textbf{WMT23/24-Biomedical}~\citep{neves-etal-2023-findings,neves-etal-2024-findings} benchmark tests reasoning within a specialized domain, demanding accurate translation of technical terminology from PubMed abstracts.

To probe more targeted reasoning abilities, we incorporate four specialized datasets. \textbf{CAMT}~\citep{yao2024benchmarking} assesses cross-cultural reasoning on expressions requiring cultural adaptation. 
\textbf{Commonsense-MT}~\citep{he-etal-2020-box} comprises subsets targeting lexical, contextless syntactic, and contextual syntactic ambiguities, each requiring commonsense reasoning.  
\textbf{RTT}~\citep{zhang-etal-2023-understanding} evaluates constrained reasoning by requiring models to correctly translate specific terminology under highly constrained conditions. 
Lastly, \textbf{RAGTrans}~\citep{wang2024retrieval} examines a model's capacity to reason over and integrate retrieved external evidence into its translation. 
Collectively, these benchmarks provide a rigorous and multifaceted framework for analyzing the effects of scaling on translation reasoning.

\begin{table*}[t]
\centering
\resizebox{\textwidth}{!}{
\begin{tabular}{l l l l l l}
\toprule
\textbf{Domain} & \textbf{Datasets} & \textbf{Granularity} & \textbf{Languages} & \textbf{Language Pair(s)} & \textbf{Sample Size} \\
\midrule
\multirow{3}{*}{Literature} 
    & WMT24-Literary~\citep{wang-etal-2024-findings} & Document-level & ZH, DE, RU & 3 & 43 \\
    & MetaphorTrans~\citep{wang-etal-2025-drt} & Sentence-level & ZH, EN & 1 & 2000 \\
    & LitEval-Corpus~\citep{zhang-etal-2025-good} & Paragraph-level & ZH, EN, DE & 4 & 187 \\
\midrule
\multirow{2}{*}{Biomedical} 
    & WMT24-Biomedical~\citep{neves-etal-2024-findings} & Document-level & \multirow{2}{*}{EN, DE, ES, FR, IT, PT, RU} & \multirow{2}{*}{12} & 600 \\
    & WMT23-Biomedical~\citep{neves-etal-2023-findings} & Document-level &  &  & 585 \\
\midrule
\multirow{1}{*}{Culture} 
    & CAMT~\citep{yao2024benchmarking} & Sentence-level & EN, ES, FR, HI, TA, TE, ZH & 7 & 6948 \\
\midrule
\multirow{3}{*}{Commonsense} 
    & Commonsense-MT~\citep{he-etal-2020-box} (Lexical Ambiguity) & Sentence-level & \multirow{3}{*}{ZH, EN} & \multirow{3}{*}{1} & 400 \\
    & Commonsense-MT (Contextless Syntactic Ambiguity) & Sentence-level &  &  & 450 \\
    & Commonsense-MT (Contextual Syntactic Ambiguity) & Sentence-level &  &  & 350 \\
\midrule
\multirow{1}{*}{Terminology} 
    & RTT~\citep{zhang-etal-2023-understanding} & Sentence-level & EN, DE & 2 & 100 \\
\midrule
\multirow{1}{*}{Misc.} 
    & RAGTrans~\citep{wang2024retrieval} & Sentence-level & ZH, EN & 1 & 1999 \\
\bottomrule
\end{tabular}
}
\caption{Overview of the MT benchmarks used in our evaluation.}
\label{tab:datasets}
\end{table*}

\subsection{Models}
Our evaluation encompasses 12 RMs, including 11 open-source models from three distinct families and one proprietary model for comparison.
The open-source models investigated are as follows:
\begin{itemize}[nolistsep,noitemsep]
    \item \textbf{Qwen-3}: Six models from this family were selected, with parameter sizes of 0.6B, 1.7B, 4B, 8B, 14B, and 32B~\citep{yang2025qwen3technicalreport}.
    These models are hybrid reasoning LLMs that support seamless switching between a standard generation mode and a deliberative reasoning mode.
    \item \textbf{Cogito}: Two models, sized 3B and 8B, were included~\citep{deepcogito_cogito_v1_preview}.
    Cogito-3B and Cogito-8B are trained on top of Llama-3.2-3B and Llama-3.1-8B~\citep{grattafiori2024llama3herdmodels}, respectively, and similarly implement hybrid reasoning capabilities with controllable switching between generation and reasoning modes.
    \item \textbf{DRT}: Three models from this family were evaluated. These models are fine-tuned from existing LLMs using the training set of MetaphorTrans~\citep{wang-etal-2025-drt}. Specifically, DRT-7B, DRT-8B, and DRT-14B are built upon Qwen2.5-7B-Instruct~\citep{qwen2024qwen2}, Llama-3.1-8B-Instruct, and Qwen2.5-14B-Instruct respectively.
\end{itemize}

In addition to the open-source models, we included the proprietary model \textbf{Grok-3-Mini}.
This model provides a tunable \texttt{reasoning\_effort} parameter (equivalent to reasoning budget but can only be set to \texttt{low} or \texttt{high}) to control the amount of deliberation performed prior to generating a response.

\subsection{Evaluation Metrics}
We assess translation quality using a suite of automatic metrics, encompassing both reference-based and reference-free approaches, alongside a specialized LLM-based judge for literary texts.

\paragraph{COMET Metrics.}
For a standardized assessment, we employ two variants from the COMET framework~\citep{rei-etal-2020-comet}: the reference-based \textbf{COMET-22}~\citep{rei-etal-2022-comet} and the reference-free \textbf{COMETKiwi-22}~\citep{rei-etal-2022-cometkiwi}.

\paragraph{LLM as Judge.}
For LLM-based evaluation, we employ \texttt{Gemini-2.0-Flash}. 
We first define two general-purpose metrics, \textbf{Gemini Reference-Based (GRB)} and \textbf{Gemini Reference-Free (GRF)}, which provide a quality score on a 0-100 scale. Furthermore, for the specific challenges of literary translation, we follow~\citet{wang-etal-2025-drt} and apply the \textbf{Gemini Evaluation with Anchors (GEA)} metric exclusively to the three literary benchmarks. 
This specialized metric assesses nuances like style and expressiveness, and we collect scores at two levels of granularity: GEA$_{100} \in [0,100]$ and GEA$_5 \in {1,\dots,5}$.
The evaluation prompts are adapted from~\citet{kocmi-federmann-2023-large} and ~\citet{wang-etal-2025-drt}, are illustrated in Appendix~\ref{sec:prompts}.

\subsection{Budget Forcing}
\label{sec:budget_forcing}
We regulate test-time reasoning with a logits processor that enforces a \emph{thinking-token budget} inside a \texttt{<think>\ldots</think>} span.
While in this span, the processor counts tokens, softly encourages closure near 95\% of the budget by upweighting newline and \texttt{</think>}, then deterministically emits a newline (penultimate step) and \texttt{</think>} (final step) at the budget limit before continuing normal answer decoding.

Conversely, to probe extrapolation, we optionally insert a single ``\texttt{wait}'' token if the model attempts to stop: specifically, after at least 5 thinking tokens, if \texttt{</think>} is the next token from the output of the argmax function and the budget is not yet exhausted, we override the next token to ``\texttt{wait}'' once and resume unconstrained decoding.
We insert ``\texttt{wait}'' at most once.

\subsection{Post-Editing}
Post-editing (or self-correction) involves two-stage translation, i.e., stage one of direct translation and stage two that corrects or post-edits the direct translation, which enables models to review and refine their own outputs~\citep{feng-etal-2025-tear, wang-etal-2024-taste, liimproving}.
We explore two prompting strategies to guide this self-correction process, with full details provided in Appendix~\ref{sec:post-editing-prompts}. 
The first is a standard PE prompt, which we term ``No QS'' (No Quality Score). 
It provides the model with only the source text and its own draft translation to be refined. 
The second is an enhanced prompt, ``QS'' (with Quality Score), which additionally includes a numerical quality score of the draft, calculated as the average of the GRB and GRF scores from the initial translation.
This provides the model with an explicit signal about the quality of the translation it needs to correct, potentially guiding a more targeted reasoning process.

\begin{figure*}[ht!]
    \centering
    \includegraphics[width=\textwidth]{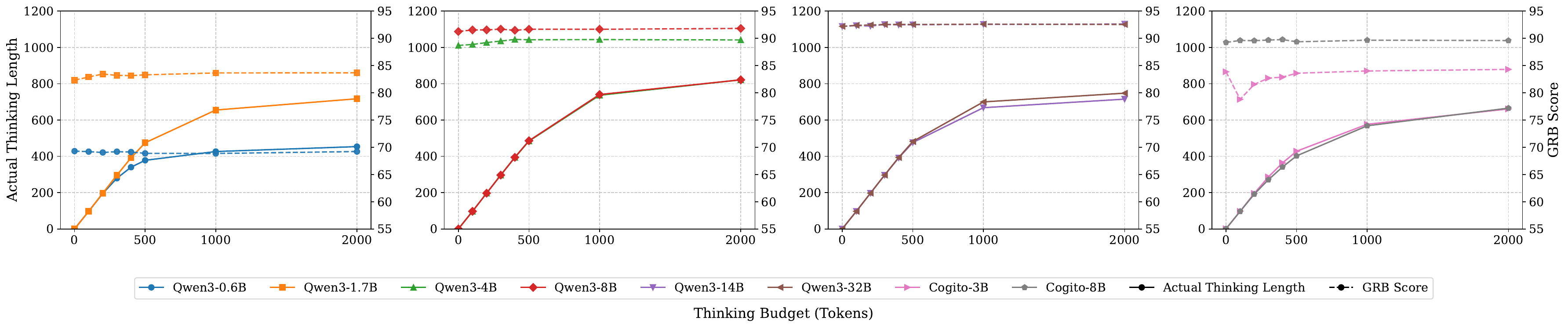}
    \caption{Average GRB scores and average actual thinking tokens of Qwen-3 and Cogito models across all datasets with varying thinking budgets.}
    \label{fig:qwen_and_cogito_grb_average}
\end{figure*}

\section{Results and Analysis}

\subsection{Effectiveness of Test-Time Scaling}

\paragraph{General-Purpose Models Show Limited Gains from Increased Budget.}
Our initial investigation focused on the efficacy of test-time scaling for general-purpose RMs, including the Qwen-3 and Cogito families, as well as the proprietary Grok-3-Mini model.
These models were evaluated ``out-of-the-box" without any fine-tuning on our benchmark datasets.
Figure~\ref{fig:qwen_and_cogito_grb_average} plots the average GRB scores (right axis) alongside the actual thinking tokens generated (left axis) for the Qwen-3 and Cogito model series.
After a small initial performance gain when moving from a zero budget to a minimal budget (e.g., 100 tokens), the models' performance curves almost completely plateau in most cases.
A critical observation emerges from the thinking-token curves: despite being allocated budgets up to 2000 tokens, most general-purpose models fail to utilize this capacity.
The actual reasoning length typically saturates around 600 tokens.
This suggests that these models reach a ``reasoning ceiling'' where they autonomously terminate the deliberation process, indicating that \textit{simply allocating more computational steps does not enable the models to produce more refined or accurate translations when they lack specific task-related knowledge.}

\begin{figure}[t!]
    \centering
    \begin{subfigure}{\columnwidth}
        \centering
        \includegraphics[width=\linewidth]{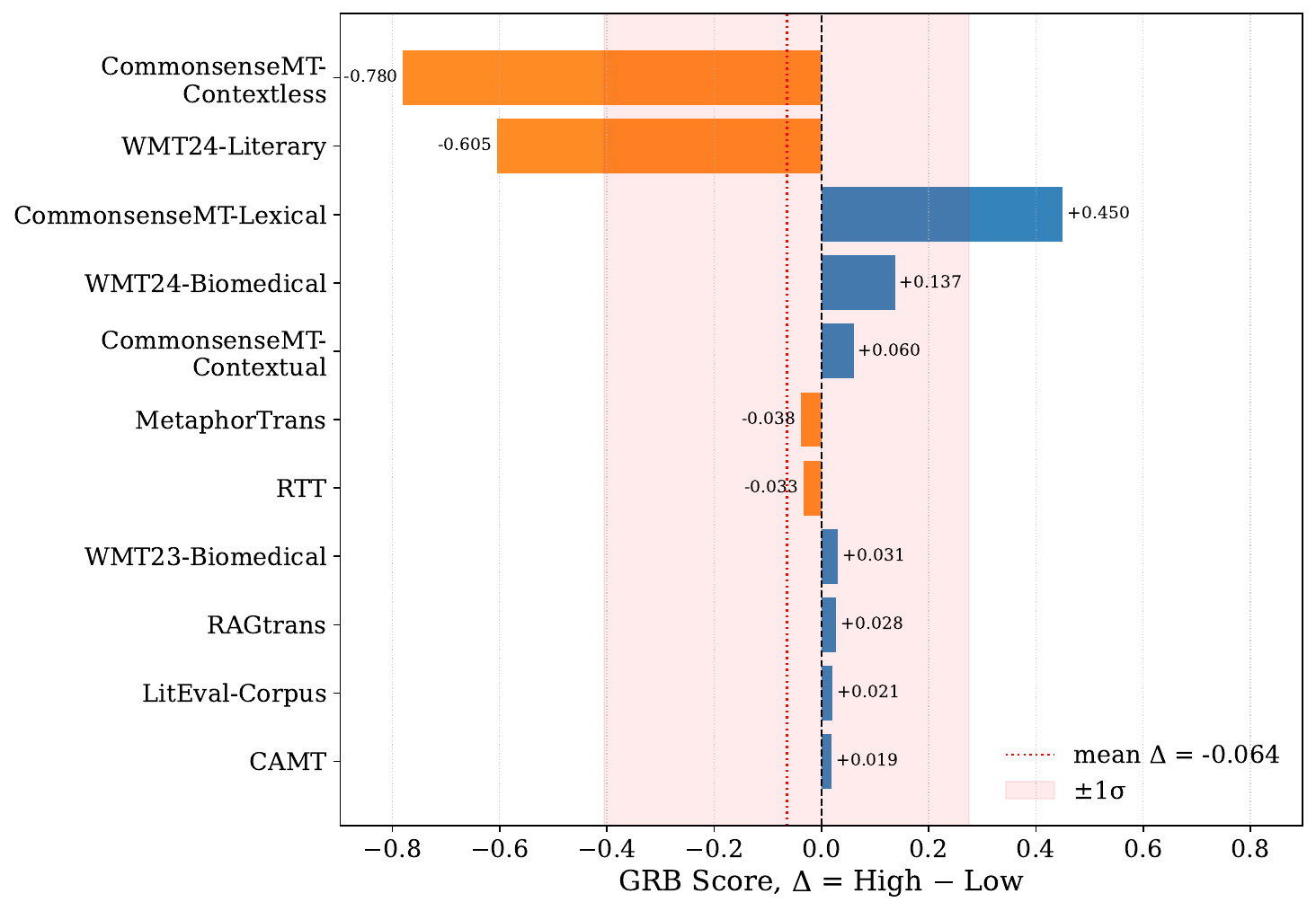}
        \caption{}
        \label{fig:grok_grb_delta_bar}
    \end{subfigure}

    \vspace{0.5em}

    \begin{subfigure}{\columnwidth}
        \centering
        \includegraphics[width=\linewidth]{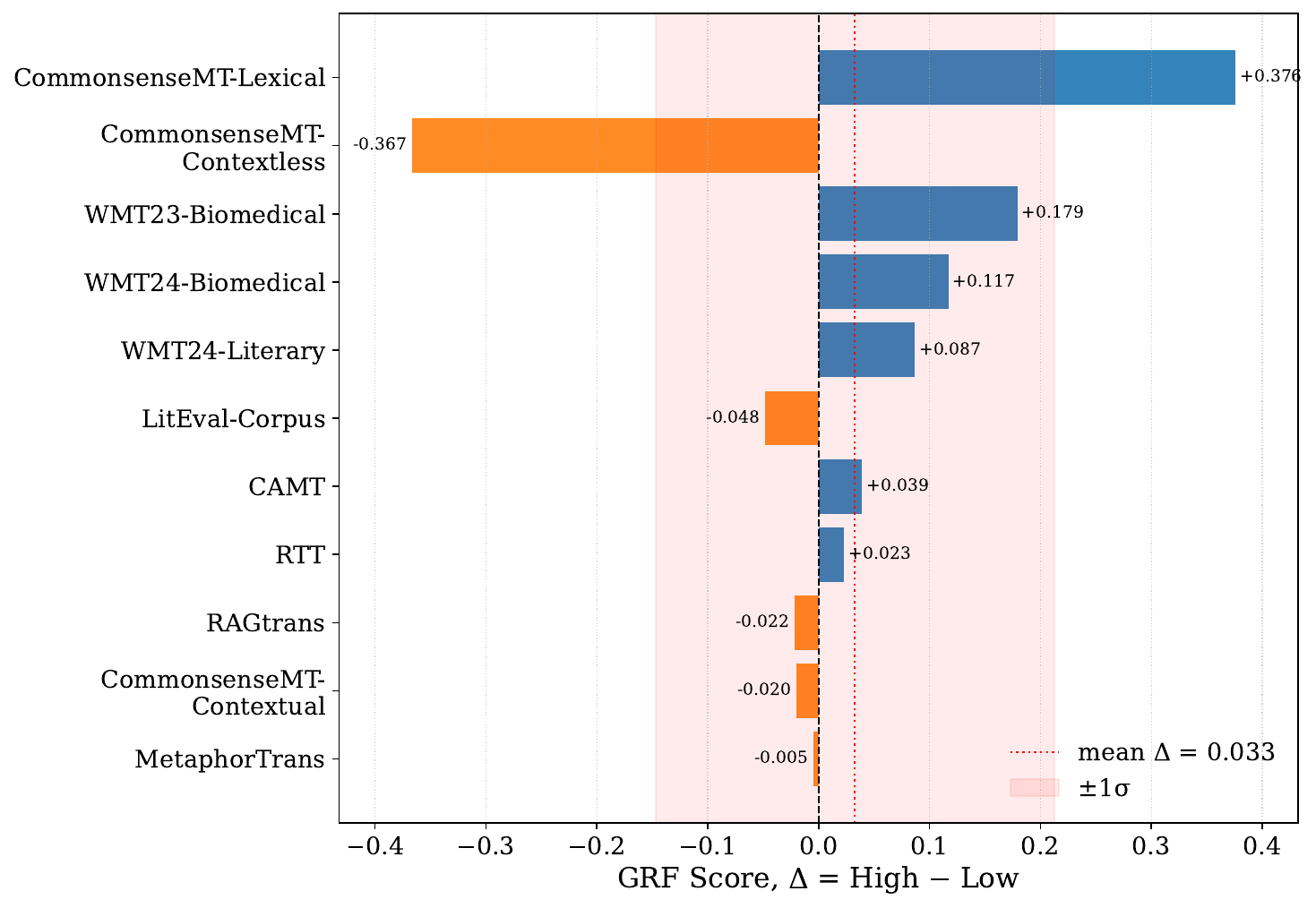}
        \caption{}
        \label{fig:grok_grf_delta_bar}
    \end{subfigure}

    \caption{Performance of Grok-3-mini across tasks, showing the difference between high- and low-effort reasoning. 
    Subfigure (a) reports results under the GRB metric, and (b) shows results under GRF.}
    \label{fig:drt_gea_combined}
\end{figure}

This conclusion is further corroborated and nuanced by our analysis of the Grok-3-Mini model. 
We analyzed its performance using both reference-based (GRB) and reference-free (GRF) metrics, visualized in Figure~\ref{fig:grok_grb_delta_bar} and Figure~\ref{fig:grok_grf_delta_bar}, respectively. 
Both metrics reveal a highly inconsistent, dataset-dependent impact. 
For instance, while higher effort improves scores on CommonsenseMT-Lexical across both GRB (+0.450) and GRF (+0.376), it significantly degrades performance on CommonsenseMT-Contextless in both cases (-0.780 for GRB, -0.367 for GRF). 
Crucially, the average effect across all datasets is negligible and even flips its sign depending on the metric: the mean GRB delta is a slightly negative -0.064, while the mean GRF delta is a slightly positive +0.033. 
Given that both scores are on a 100-point scale, these near-zero average changes underscore that the potential benefits and drawbacks of increased computational effort effectively cancel each other out, leading to no reliable overall improvement.

Therefore, our analyses of both open-source models with varying budgets and a proprietary model with different effort levels converge on a single conclusion: \textit{for general-purpose LLMs without specific in-domain training, test-time scaling is not a robust strategy for enhancing machine translation performance.}

\begin{figure}[t]
    \centering
    \includegraphics[width=\columnwidth]{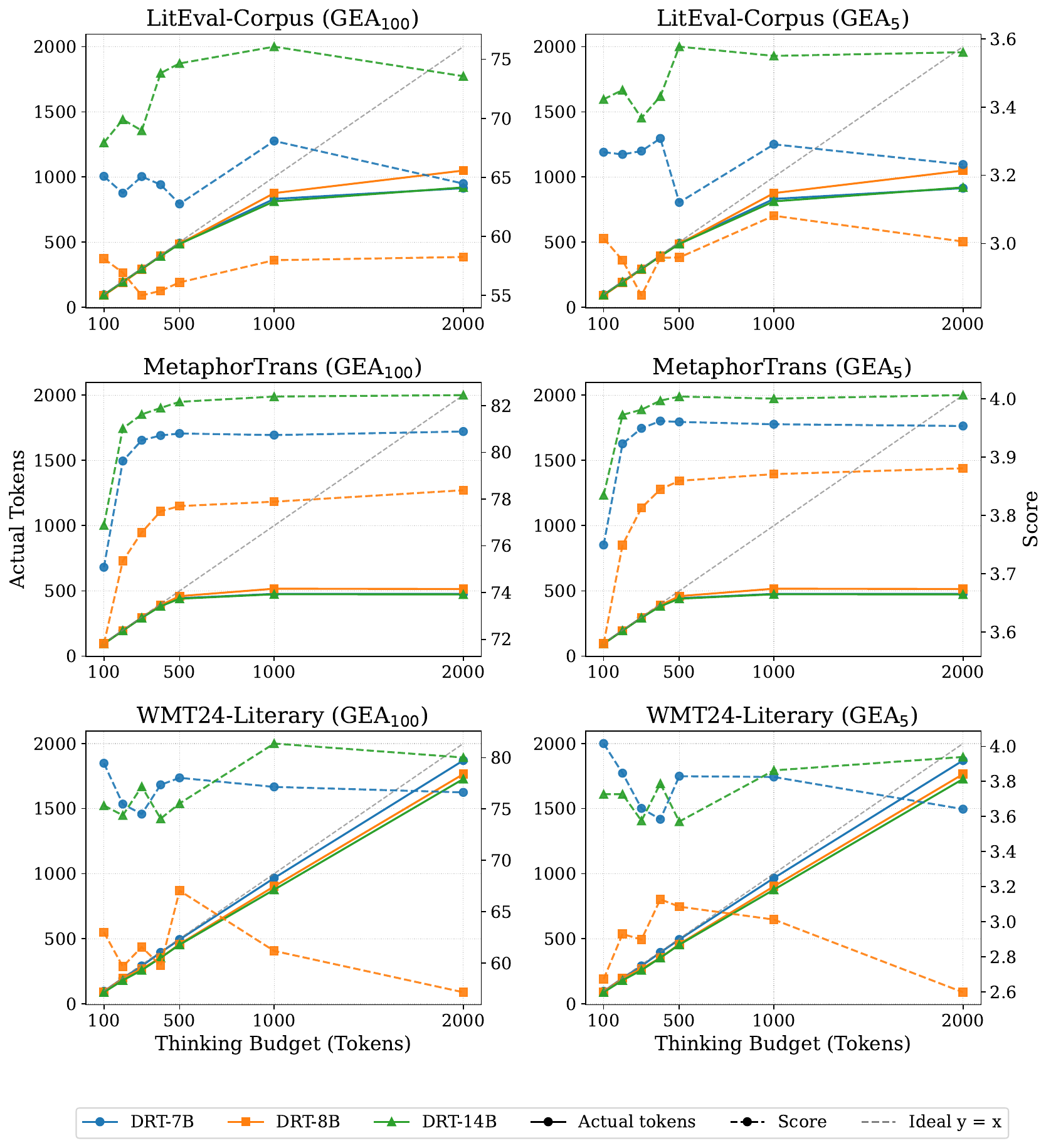}
    \caption{Performance (dashed lines, right axis) and actual generated thinking tokens (solid lines, left axis) of DRT models across 3 literary translation tasks. 
    }
    \label{fig:drt_budget_vs_actual_combined}
\end{figure}

\paragraph{In-Domain Fine-Tuning Unlocks the Benefit of Test-Time Scaling.}
In contrast to the general-purpose models, the DRT models reveal that the effectiveness of test-time scaling is highly contingent on domain-specific training, which appears to create an efficient alignment between reasoning effort and performance. 
These models were fine-tuned on the training set of MetaphorTrans (in-domain), while LitEval-Corpus and WMT24-Literary serve as related but out-of-domain literary benchmarks.
Figure~\ref{fig:drt_budget_vs_actual_combined} visualizes the translation quality (GEA score, right y-axis) and the actual number of thinking tokens generated by the models (left y-axis).

On the in-domain MetaphorTrans task, we observe a clear and consistent positive correlation between the thinking budget and translation performance.
We consider the model's natural stopping point, where the `Actual tokens' curve plateaus around 500 tokens, as the realistic baseline.
As the thinking budget increases from 100 to this limit, both the number of generated tokens and the GEA scores steadily rise.
This monotonic improvement indicates that the model is performing valid, necessary reasoning steps.

However, beyond a 500-token budget, a critical pattern emerges: the models stop generating more thinking tokens, and concurrently, their performance plateaus.
We hypothesize that the fine-tuning successfully aligns the model's reasoning behavior with the effective reasoning boundary of the task.

This efficient alignment vanishes on the other out-of-domain literary translation tasks. 
The most striking counterexample is the document-level WMT24-Literary task.
Here, the actual thinking tokens continue to scale almost linearly with the budget, indicating the models are using the provided extra capacity to ``think" longer.\footnote{We attribute this distinct behavior to text granularity: unlike the out-of-domain paragraph-level tasks (LitEval-Corpus) where limited context leads models to exhaust reasoning paths and saturate early, the extensive context in document-level translation allows models to continuously expand their reasoning loops to consume the available budget.}
Yet, this extended reasoning does not translate into better performance; the GEA scores remain erratic and show no consistent improvement.
This disconnect suggests the models are engaged in unproductive or unfocused reasoning, ``spinning their wheels" without the specialized knowledge required for this different type of literary translation. 
This dichotomy underscores our central argument: \textit{test-time scaling is most effective when fine-tuning has equipped a model with not only domain-specific knowledge but also an efficient strategy for how and when to apply it.}





\subsection{Forced Extrapolation of Reasoning Degrades Performance}
Building on the observation that models possess a natural reasoning length, we next address RQ2: Does performance improve if we force a model to think longer? 
We test this by applying the ``\texttt{wait}'' token extrapolation method, detailed in Section~\ref{sec:budget_forcing}, to prompt continued reasoning when the models are about to stop the thinking process.

The results, averaged across all datasets for the Qwen-3 and Cogito models, are presented in Table~\ref{tab:wait_token}.
The findings are unambiguous. 
First, as evidenced by the ``Thinking Length'' columns, the intervention was effective in its primary goal: it consistently and significantly extended the models' reasoning chains, often by over 100-200 tokens. 

However, this artificially prolonged reasoning process did not translate into better translations. 
In fact, it was overwhelmingly detrimental. 
Across all four metrics (COMET, COMETKiwi, GRB, and GRF), the ``\texttt{wait}'' token intervention consistently reduces performance: under both the 1000- and 2000-token budgets, 55 of the 64 metric scores across eight models dropped after forced extrapolation.
While there are a few isolated instances of negligible score increases in one metric (e.g., Qwen3-4B on GRB with a 2000-token budget), these are exceptions that are contradicted by decreases in other metrics for the same model.

This leads to a clear conclusion: a model's decision to terminate its reasoning chain is a meaningful signal. 
It indicates that the model has reached what it considers to be a sufficient state of deliberation for the given task. 
\textit{Forcing the model to continue to reason beyond its own stopping point appears to introduce noise, repetition, or less relevant reasoning steps, which ultimately harms the quality of the final translation.} 
In short, we find that forced extrapolation is a counterproductive strategy for improving translation quality.

\begin{table*}[t]
\centering
\resizebox{\textwidth}{!}{%
\begin{tabular}{l c
                c c
                c c
                c c
                c c
                c c}
\toprule
\multirow{2}{*}{Model} & \multirow{2}{*}{Budget} &
\multicolumn{2}{c}{Thinking Length} &
\multicolumn{2}{c}{COMET} &
\multicolumn{2}{c}{COMETKiwi} &
\multicolumn{2}{c}{GRB} &
\multicolumn{2}{c}{GRF} \\
\cmidrule(lr){3-4}
\cmidrule(lr){5-6}
\cmidrule(lr){7-8}
\cmidrule(lr){9-10}
\cmidrule(lr){11-12}
 & &
{Before} & {After} &
{Before} & {After} &
{Before} & {After} &
{Before} & {After} &
{Before} & {After} \\
\midrule
\multirow{2}{*}{Qwen3-0.6B} & 1000 & 426 & 519 & \textbf{0.6959} & 0.6904 & \textbf{0.6155} & 0.6026 & \textbf{68.8577} & 68.6333 & \textbf{67.2944} & 67.1203 \\
            & 
            \cellcolor{gray!15}2000 & 
            \cellcolor{gray!15}454 & 
            \cellcolor{gray!15}556 & 
            \cellcolor{gray!15}\textbf{0.6895} & 
            \cellcolor{gray!15}0.6894 & 
            \cellcolor{gray!15}\textbf{0.6087} & 
            \cellcolor{gray!15}0.6016 & 
            \cellcolor{gray!15}\textbf{69.2139} & 
            \cellcolor{gray!15}69.0237 & 
            \cellcolor{gray!15}\textbf{67.6465} & 
            \cellcolor{gray!15}67.3326 \\
\multirow{2}{*}{Qwen3-1.7B} & 1000 & 655 & 820 & \textbf{0.7687} & 0.7475 & \textbf{0.6851} & 0.6543 & \textbf{83.6481} & 83.5185 & \textbf{83.1170} & 82.7297 \\
            & 
            \cellcolor{gray!15}2000 & 
            \cellcolor{gray!15}717 & 
            \cellcolor{gray!15}987 & 
            \cellcolor{gray!15}\textbf{0.7645} & 
            \cellcolor{gray!15}0.7496 & 
            \cellcolor{gray!15}\textbf{0.6762} & 
            \cellcolor{gray!15}0.6586 & 
            \cellcolor{gray!15}\textbf{83.6647} & 
            \cellcolor{gray!15}83.4715 & 
            \cellcolor{gray!15}\textbf{83.1076} & 
            \cellcolor{gray!15}82.7472 \\
\multirow{2}{*}{Qwen3-4B} & 1000 & 737 & 873 & \textbf{0.7914} & 0.7738 & \textbf{0.7092} & 0.6835 & \textbf{89.7781} & 89.5598 & \textbf{89.7522} & 89.5203 \\
            & 
            \cellcolor{gray!15}2000 & 
            \cellcolor{gray!15}822 & 
            \cellcolor{gray!15}1098 & 
            \cellcolor{gray!15}\textbf{0.7871} & 
            \cellcolor{gray!15}0.7784 & 
            \cellcolor{gray!15}\textbf{0.7005} & 
            \cellcolor{gray!15}0.6862 & 
            \cellcolor{gray!15}89.7023 & 
            \cellcolor{gray!15}\textbf{90.0098} & 
            \cellcolor{gray!15}89.6536 & 
            \cellcolor{gray!15}\textbf{89.7275} \\
\multirow{2}{*}{Qwen3-8B} & 1000 & 741 & 878 & \textbf{0.7979} & 0.7865 & \textbf{0.7180} & 0.6954 & 91.6706 & \textbf{91.7601} & \textbf{91.8085} & 91.7027 \\
            & 
            \cellcolor{gray!15}2000 & 
            \cellcolor{gray!15}821 & 
            \cellcolor{gray!15}1117 & 
            \cellcolor{gray!15}\textbf{0.7965} & 
            \cellcolor{gray!15}0.7860 & 
            \cellcolor{gray!15}\textbf{0.7118} & 
            \cellcolor{gray!15}0.6951 & 
            \cellcolor{gray!15}\textbf{91.8186} & 
            \cellcolor{gray!15}91.6111 & 
            \cellcolor{gray!15}91.8123 & 
            \cellcolor{gray!15}\textbf{91.9008} \\
\multirow{2}{*}{Qwen3-14B} & 1000 & 668 & 812 & \textbf{0.7992} & 0.7924 & \textbf{0.7184} & 0.7022 & 92.5609 & \textbf{92.5677} & \textbf{92.5242} & 92.5148 \\
            & 
            \cellcolor{gray!15}2000 & 
            \cellcolor{gray!15}715 & 
            \cellcolor{gray!15}955 & 
            \cellcolor{gray!15}\textbf{0.7966} & 
            \cellcolor{gray!15}0.7908 & 
            \cellcolor{gray!15}\textbf{0.7125} & 
            \cellcolor{gray!15}0.7020 & 
            \cellcolor{gray!15}\textbf{92.6259} & 
            \cellcolor{gray!15}92.5421 & 
            \cellcolor{gray!15}\textbf{92.5617} & 
            \cellcolor{gray!15}92.5392 \\
\multirow{2}{*}{Qwen3-32B} & 1000 & 700 & 835 & \textbf{0.8025} & 0.7828 & \textbf{0.7233} & 0.6957 & \textbf{92.6026} & 92.5537 & \textbf{92.6328} & 92.5322 \\
            & 
            \cellcolor{gray!15}2000 & 
            \cellcolor{gray!15}748 & 
            \cellcolor{gray!15}992 & 
            \cellcolor{gray!15}\textbf{0.7993} & 
            \cellcolor{gray!15}0.7786 & 
            \cellcolor{gray!15}\textbf{0.7174} & 
            \cellcolor{gray!15}0.6937 & 
            \cellcolor{gray!15}92.5393 & 
            \cellcolor{gray!15}\textbf{92.6689} & 
            \cellcolor{gray!15}92.6698 & 
            \cellcolor{gray!15}\textbf{92.7696} \\
\multirow{2}{*}{Cogito-3B} & 1000 & 577 & 695 & \textbf{0.7071} & 0.7063 & \textbf{0.6313} & 0.6310 & \textbf{84.0098} & 83.8203 & \textbf{83.1502} & 82.5440 \\
            & 
            \cellcolor{gray!15}2000 & 
            \cellcolor{gray!15}661 & 
            \cellcolor{gray!15}844 & 
            \cellcolor{gray!15}0.7040 & 
            \cellcolor{gray!15}\textbf{0.7049} & 
            \cellcolor{gray!15}0.6324 & 
            \cellcolor{gray!15}\textbf{0.6331} & 
            \cellcolor{gray!15}\textbf{84.2981} & 
            \cellcolor{gray!15}83.9921 & 
            \cellcolor{gray!15}\textbf{82.4356} & 
            \cellcolor{gray!15}81.8855 \\
\multirow{2}{*}{Cogito-8B} & 1000 & 568 & 697 & \textbf{0.7678} & 0.7658 & \textbf{0.6824} & 0.6807 & \textbf{89.6573} & 89.3849 & \textbf{89.2655} & 88.8525 \\
            & 
            \cellcolor{gray!15}2000 & 
            \cellcolor{gray!15}665 & 
            \cellcolor{gray!15}882 & 
            \cellcolor{gray!15}\textbf{0.7687} & 
            \cellcolor{gray!15}\textbf{0.7687} & 
            \cellcolor{gray!15}\textbf{0.6823} & 
            \cellcolor{gray!15}0.6809 & 
            \cellcolor{gray!15}\textbf{89.5962} & 
            \cellcolor{gray!15}89.4339 & 
            \cellcolor{gray!15}\textbf{89.3591} & 
            \cellcolor{gray!15}89.2452 \\
\bottomrule
\end{tabular}%
}
\caption{Effect of forcibly inserting a ``\texttt{wait}'' token to extend the reasoning process. 
The ``Before'' columns show standard generation, while ``After'' shows results from the intervention. 
}
\label{tab:wait_token}
\end{table*}

\subsection{Test-Time Scaling is Effective for Post-Editing}
\begin{figure*}[htbp]
    \centering
    \begin{subfigure}{0.48\textwidth}
        \centering
        \includegraphics[width=\linewidth]{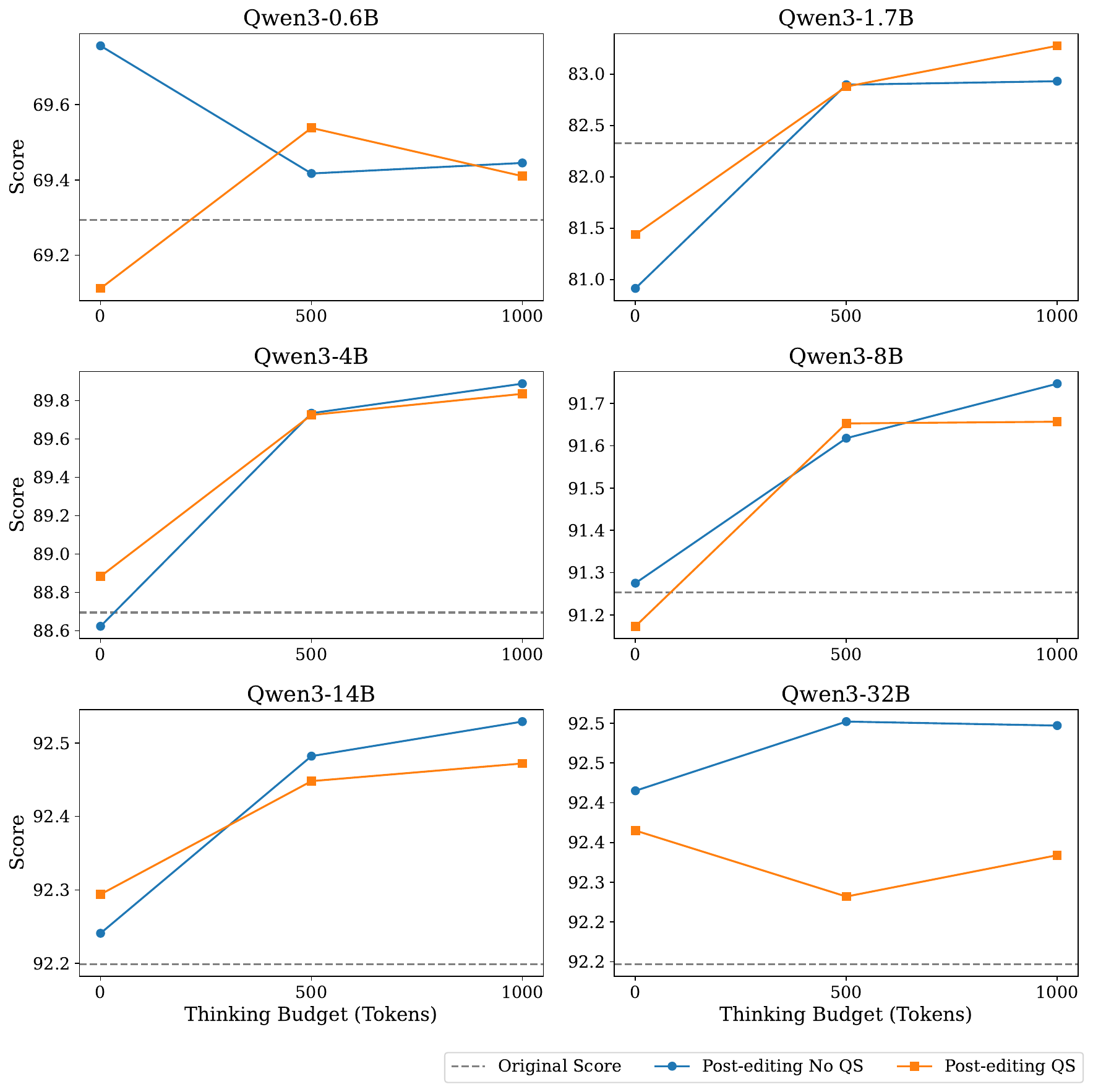}
        \caption{GRB scores for post-editing across Qwen-3 models.}
        \label{fig:post_editing_compare_by_model_grb}
    \end{subfigure}
    \hfill
    \begin{subfigure}{0.48\textwidth}
        \centering
        \includegraphics[width=\linewidth]{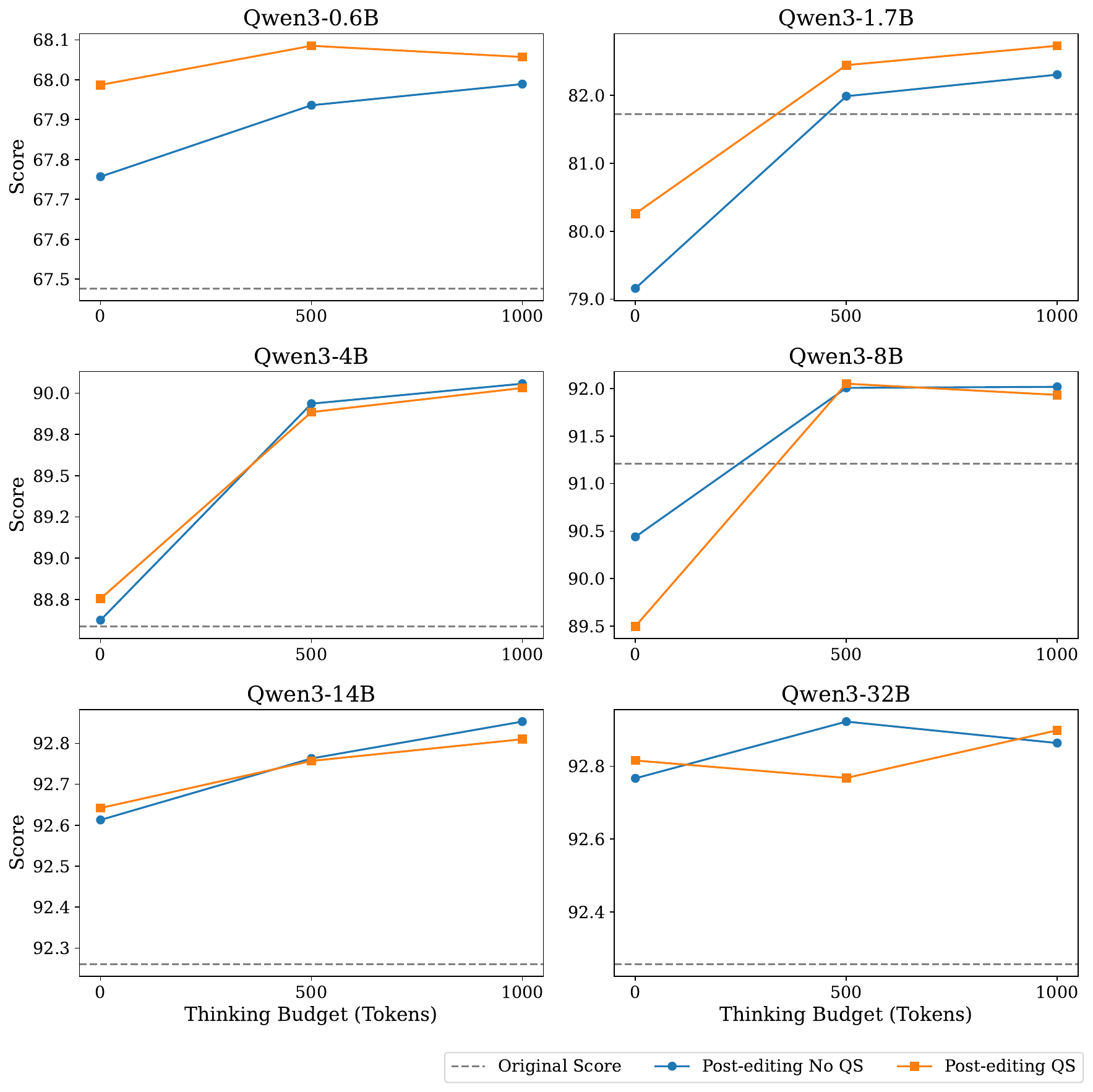}
        \caption{GRF scores for post-editing across Qwen-3 models.}
        \label{fig:post_editing_compare_by_model_grf}
    \end{subfigure}

    \caption{Effectiveness of test-time scaling in post-editing scenario.}
    \label{fig:post_editing_compare_by_model}
\end{figure*}

Finally, we investigate RQ3 by evaluating the effectiveness of test-time scaling in a self-correction post-editing scenario.
For this task, we define the baseline as the translation generated by each Qwen-3 model with a zero thinking budget in our prior experiments.
Subsequently, we task the same model with refining its own translation, applying thinking budgets of 0, 500, and 1000 tokens.
The detailed results are presented in Table~\ref{tab:post-editing-grb} and Table~\ref{tab:post-editing-grf}, with trends visualized in Figure~\ref{fig:post_editing_compare_by_model}.


In a striking contrast to its ineffectiveness in direct translation, \textit{test-time scaling proves to be a highly effective strategy for post-editing, reliably elevating translation quality above the original baseline for most models.}
The effect is most pronounced for mid-sized models, as shown in both the GRB (Figure~\ref{fig:post_editing_compare_by_model_grb}) and GRF (Figure~\ref{fig:post_editing_compare_by_model_grf}) plots.
For models in the 1.7B to 14B parameter range, applying post-editing with a zero budget often yields results that are similar to or worse than the original translation.
Increasing the budget to 500 or 1000 tokens consistently pushes performance significantly above this baseline, demonstrating that a thinking budget is crucial for turning self-correction into a reliably beneficial process.

However, this scaling trend does not hold for the extremes of the model family. 
The smallest model, Qwen3-0.6B, displays erratic behavior, with its performance fluctuating without clear improvement as the budget increases. 
Conversely, the largest model, Qwen3-32B, already surpasses the baseline with a zero-budget correction, and additional thinking time provides no further gains, suggesting it performs near its peak without extended deliberation.

A comparison of the two prompting strategies further highlights the importance of the thinking budget. 
At a zero-token budget, the ``QS'' prompt (with quality score) generally underperforms the ``No QS'' prompt. 
However, once the budget is increased to 500 or 1000 tokens, their performances converge and become nearly indistinguishable.
This demonstrates that while prompting strategy matters, it is the allocation of a computational budget that is the key factor for post-editing to reliably improve upon the initial translation.





\section{Related Work}
\paragraph{Machine Translation with Large Language Models}
The application of LLMs to machine translation has witnessed a paradigm shift.
Foundational research demonstrated that general-purpose models, such as GPT-3, possess remarkable few-shot translation capabilities, challenging traditional supervised systems~\citep{brown2020language}.
Subsequent empirical studies systematically evaluated these capabilities, revealing that while LLMs excel in high-resource languages and specific domains, they often require careful adaptation to match state-of-the-art baselines~\citep{hendy2023good, zhu-etal-2024-multilingual}.
To address these limitations, researchers have developed sophisticated in-context learning strategies \citep{agrawal-etal-2023-context} and diverse prompting techniques~\citep{vilar-etal-2023-prompting}.
Notably, recent work by~\citet{wu-etal-2025-please} highlights the efficacy of simple re-translation prompts over complex reasoning in general LLMs, suggesting that iterative refinement can significantly boost performance without elaborate chains of thought.
Concurrently, the field is moving towards open-weight models that balance broad multilingual competence with translation specialization.
Seed-X~\citep{cheng2025seedxbuildingstrongmultilingual} introduces a 7B-parameter open-source family of translation-oriented LLMs trained on large-scale monolingual and bilingual corpora across 28 languages, achieving performance competitive with closed-source systems such as GPT-4o~\citep{hurst2024gpt} and Gemini-2.5~\citep{comanici2025gemini}.
Similarly, Hunyuan-MT~\citep{zheng2025hunyuan} develops 7B-parameter models: Hunyuan-MT and Hunyuan-MT-Chimera, the latter integrates multiple outputs under a ``slow thinking'' paradigm to yield higher-quality translations, and ranks first in the WMT2025 shared task across 30 of 31 directions~\citep{kocmi-etal-2025-findings}.
Tower+~\citep{rei2025tower+} addresses the trade-off between translation specialization and general-purpose ability by combining continued pretraining, supervised fine-tuning, preference optimization, and reinforcement learning with verifiable rewards.

\paragraph{Machine Translation with Reasoning Models}
Recent research has explored how RMs can be adapted to MT, particularly in linguistically and culturally challenging domains.
\citet{wang-etal-2025-drt} propose Deep Reasoning Translation (DRT), which leverages long chain-of-thought (CoT) reasoning within a multi-agent framework to tackle similes and metaphors in English–Chinese literary translation.
The resulting models surpass standard LLMs by synthesizing long-thought training data.
Building on this, DeepTrans~\citep{wang2025deep} applies reinforcement learning (RL) with carefully designed reward functions targeting both translation fidelity and reasoning quality, showing that RL without labeled pairs can significantly boost performance. 
ExTrans~\citep{wang2025extrans} complements this direction with an exemplar-enhanced RL approach that employs a stronger RM (DeepSeek-R1) as a reward reference. 
ExTrans achieves state-of-the-art results in English–Chinese literary MT, and its multilingual extension (mExTrans) scales effectively to 90 directions with lightweight reward modeling. 
Beyond literary MT, R1-T1~\citep{he2025r1t1fullyincentivizingtranslation} generalizes reasoning-based MT by modeling six CoT templates inspired by human translator strategies. 
Through RL, it enables self-evolving reasoning trajectories, improving performance across diverse domains and low-resource languages.

\paragraph{Test-Time Scaling}
Recent studies have examined the potential of test-time scaling.
\citet{tan2025investigating} propose a best-of-$N$ reranking framework where multiple translation candidates are generated and the best one is selected using a quality estimation model.
They show that smaller models can, through TTS, match or even surpass larger model. 
For example, a 14B model with $N\!\approx\!8$ achieves parity with a 72B model at $N=1$ while requiring substantially less GPU memory. 
Beyond MT, \citet{son2025linguistic} analyze TTS in multilingual mathematical reasoning, finding that outcome and process reward modeling, as well as budget forcing, yield notable gains in English but limited improvements across 55 languages, highlighting cross-lingual fragility. 
\citet{tran2025scaling} study low-resource reasoning tasks and propose English-pivoted CoT generation, where reasoning occurs in English before producing final answers in the target language, yielding substantial accuracy improvements. 
\citet{yong2025crosslingual} further study cross-lingual reasoning with English-centric RMs, finding that scaling inference budgets with long CoTs improves multilingual mathematical reasoning and even allows smaller models to outperform larger baselines, but also highlighting language-mixing behaviors and weaker generalization to cultural commonsense domains.

\section{Conclusion}
In this work, we systematically explored the application of test-time scaling (TTS) to reasoning models (RMs) for machine translation (MT), addressing three core research questions through extensive experiments across diverse benchmarks, models, and evaluation metrics.

Our findings reveal that TTS offers limited value for general-purpose RMs in direct translation tasks, where performance quickly plateaus after minimal initial gains, underscoring that additional inference-time computation alone cannot compensate for a lack of task-aligned reasoning strategies. 
In contrast, domain-specific fine-tuning emerges as a pivotal enabler, allowing TTS to yield consistent improvements on in-domain tasks until models reach their natural reasoning depth, beyond which further scaling provides no benefit. 
This highlights an emergent efficiency in fine-tuned models, where optimal deliberation aligns with task demands, though such alignment erodes out-of-domain. 
Furthermore, forcibly extending reasoning via interventions like ``\texttt{wait}'' tokens consistently degrades quality, emphasizing the importance of respecting a model's intrinsic stopping points. 
Finally, TTS proves particularly potent in post-editing scenarios, transforming self-correction into a reliable mechanism for refining initial drafts, especially for mid-sized models when paired with adequate budgets.

The implications of this work are twofold.  
First, for practitioners, simply allocating more inference compute to general-purpose models is an inefficient path to better translations. 
Instead, resources are better invested in targeted fine-tuning, which aligns the model's reasoning capabilities with specific task demands.  
Second, our results suggest that the most promising application of TTS in MT is not in direct, single-pass translation but in multi-stage workflows, such as a rapid initial draft followed by a more deliberate, computationally-intensive self-correction phase.  
Future work could explore more dynamic budget allocation strategies and extend to hybrid TTS approaches integrated with external tools like retrieval-augmented generation.

\section*{Limitations}
While our study provides a comprehensive analysis of test-time scaling in machine translation, we acknowledge several limitations that frame the scope of our conclusions and suggest avenues for future research.

First, our investigation, while encompassing 12 different models, is primarily focused on open-source RM families and a single, smaller proprietary model. 
The performance characteristics and scaling behaviors of the largest, state-of-the-art proprietary models (e.g., Gemini-2.5-Pro) may differ from our observations. 
Furthermore, the linguistic diversity of our benchmarks is largely centered around English or Chinese as either a source or target language. 
Consequently, our findings on the effectiveness of TTS, particularly the interplay with fine-tuning, may not generalize directly to low-resource languages where the reasoning challenges could be substantially different.

Second, our evaluation methodology relies exclusively on automatic and LLM-based metrics. 
Although we employed a suite of reference-based, reference-free, and specialized LLM-judge metrics to ensure robustness, this approach lacks the nuance of human evaluation. 
A human assessment would be invaluable for validating our findings, especially on the literary and cultural benchmarks where subtle aspects of style, tone, and appropriateness are critical and may not be fully captured by our current metrics. 
The potential biases inherent in LLM-as-a-judge frameworks also represent a confounding factor.

Finally, our study implements test-time scaling through a specific budget-forcing mechanism and a simple ``\texttt{wait}'' token intervention for extrapolation. 
Other methods for encouraging or extending deliberation, such as alternative prompting strategies or more complex reasoning frameworks, were not explored and could yield different outcomes. 
Additionally, our analysis is primarily quantitative; we did not perform a qualitative analysis of the content within the models' reasoning chains. 
A deeper investigation into what the models are ``thinking'' could provide valuable insights into why performance plateaus for general-purpose models or why forced extrapolation leads to degradation.

\section*{Acknowledgments}
The work has received funding from the Digital Europe Programme under grant agreement No 101195233 (OpenEuroLLM). 
The authors wish to acknowledge CSC - IT Center for Science, Finland, and LUMI supercomputers, owned by the EuroHPC Joint Undertaking, for providing computational resources.
Shaoxiong Ji received support from the CA21167 COST action UniDive, funded by COST (European Cooperation in Science and Technology).

\bibliography{custom}

@misc{yang2025qwen3technicalreport,
      title={Qwen3 Technical Report}, 
      author={An Yang and Anfeng Li and Baosong Yang and Beichen Zhang and Binyuan Hui and Bo Zheng and Bowen Yu and Chang Gao and Chengen Huang and Chenxu Lv and Chujie Zheng and Dayiheng Liu and Fan Zhou and Fei Huang and Feng Hu and Hao Ge and Haoran Wei and Huan Lin and Jialong Tang and Jian Yang and Jianhong Tu and Jianwei Zhang and Jianxin Yang and Jiaxi Yang and Jing Zhou and Jingren Zhou and Junyang Lin and Kai Dang and Keqin Bao and Kexin Yang and Le Yu and Lianghao Deng and Mei Li and Mingfeng Xue and Mingze Li and Pei Zhang and Peng Wang and Qin Zhu and Rui Men and Ruize Gao and Shixuan Liu and Shuang Luo and Tianhao Li and Tianyi Tang and Wenbiao Yin and Xingzhang Ren and Xinyu Wang and Xinyu Zhang and Xuancheng Ren and Yang Fan and Yang Su and Yichang Zhang and Yinger Zhang and Yu Wan and Yuqiong Liu and Zekun Wang and Zeyu Cui and Zhenru Zhang and Zhipeng Zhou and Zihan Qiu},
      year={2025},
      eprint={2505.09388},
      archivePrefix={arXiv},
      primaryClass={cs.CL},
      url={https://arxiv.org/abs/2505.09388}, 
}

@misc{deepcogito_cogito_v1_preview,
  author       = {{Deep Cogito}},
  title        = {{Cogito v1 Preview Introducing IDA as a path to general superintelligence}},
  howpublished = {\url{https://www.deepcogito.com/research/cogito-v1-preview}},
  year         = {2025},
  month        = apr,
  note         = {Accessed: 2025-09-09},
}

@inproceedings{wang-etal-2025-drt,
    title = "{DRT}: Deep Reasoning Translation via Long Chain-of-Thought",
    author = "Wang, Jiaan  and
      Meng, Fandong  and
      Liang, Yunlong  and
      Zhou, Jie",
    editor = "Che, Wanxiang  and
      Nabende, Joyce  and
      Shutova, Ekaterina  and
      Pilehvar, Mohammad Taher",
    booktitle = "Findings of the Association for Computational Linguistics: ACL 2025",
    month = jul,
    year = "2025",
    address = "Vienna, Austria",
    publisher = "Association for Computational Linguistics",
    url = "https://aclanthology.org/2025.findings-acl.351/",
    doi = "10.18653/v1/2025.findings-acl.351",
    pages = "6770--6782",
    ISBN = "979-8-89176-256-5"
}

@misc{grattafiori2024llama3herdmodels,
      title={The Llama 3 Herd of Models}, 
      author={Aaron Grattafiori and Abhimanyu Dubey and Abhinav Jauhri and Abhinav Pandey and Abhishek Kadian and Ahmad Al-Dahle and Aiesha Letman and Akhil Mathur and Alan Schelten and Alex Vaughan and Amy Yang and Angela Fan and Anirudh Goyal and Anthony Hartshorn and Aobo Yang and Archi Mitra and Archie Sravankumar and Artem Korenev and Arthur Hinsvark and Arun Rao and Aston Zhang and Aurelien Rodriguez and Austen Gregerson and Ava Spataru and Baptiste Roziere and Bethany Biron and Binh Tang and Bobbie Chern and Charlotte Caucheteux and Chaya Nayak and Chloe Bi and Chris Marra and Chris McConnell and Christian Keller and Christophe Touret and Chunyang Wu and Corinne Wong and Cristian Canton Ferrer and Cyrus Nikolaidis and Damien Allonsius and Daniel Song and Danielle Pintz and Danny Livshits and Danny Wyatt and David Esiobu and Dhruv Choudhary and Dhruv Mahajan and Diego Garcia-Olano and Diego Perino and Dieuwke Hupkes and Egor Lakomkin and Ehab AlBadawy and Elina Lobanova and Emily Dinan and Eric Michael Smith and Filip Radenovic and Francisco Guzmán and Frank Zhang and Gabriel Synnaeve and Gabrielle Lee and Georgia Lewis Anderson and Govind Thattai and Graeme Nail and Gregoire Mialon and Guan Pang and Guillem Cucurell and Hailey Nguyen and Hannah Korevaar and Hu Xu and Hugo Touvron and Iliyan Zarov and Imanol Arrieta Ibarra and Isabel Kloumann and Ishan Misra and Ivan Evtimov and Jack Zhang and Jade Copet and Jaewon Lee and Jan Geffert and Jana Vranes and Jason Park and Jay Mahadeokar and Jeet Shah and Jelmer van der Linde and Jennifer Billock and Jenny Hong and Jenya Lee and Jeremy Fu and Jianfeng Chi and Jianyu Huang and Jiawen Liu and Jie Wang and Jiecao Yu and Joanna Bitton and Joe Spisak and Jongsoo Park and Joseph Rocca and Joshua Johnstun and Joshua Saxe and Junteng Jia and Kalyan Vasuden Alwala and Karthik Prasad and Kartikeya Upasani and Kate Plawiak and Ke Li and Kenneth Heafield and Kevin Stone and Khalid El-Arini and Krithika Iyer and Kshitiz Malik and Kuenley Chiu and Kunal Bhalla and Kushal Lakhotia and Lauren Rantala-Yeary and Laurens van der Maaten and Lawrence Chen and Liang Tan and Liz Jenkins and Louis Martin and Lovish Madaan and Lubo Malo and Lukas Blecher and Lukas Landzaat and Luke de Oliveira and Madeline Muzzi and Mahesh Pasupuleti and Mannat Singh and Manohar Paluri and Marcin Kardas and Maria Tsimpoukelli and Mathew Oldham and Mathieu Rita and Maya Pavlova and Melanie Kambadur and Mike Lewis and Min Si and Mitesh Kumar Singh and Mona Hassan and Naman Goyal and Narjes Torabi and Nikolay Bashlykov and Nikolay Bogoychev and Niladri Chatterji and Ning Zhang and Olivier Duchenne and Onur Çelebi and Patrick Alrassy and Pengchuan Zhang and Pengwei Li and Petar Vasic and Peter Weng and Prajjwal Bhargava and Pratik Dubal and Praveen Krishnan and Punit Singh Koura and Puxin Xu and Qing He and Qingxiao Dong and Ragavan Srinivasan and Raj Ganapathy and Ramon Calderer and Ricardo Silveira Cabral and Robert Stojnic and Roberta Raileanu and Rohan Maheswari and Rohit Girdhar and Rohit Patel and Romain Sauvestre and Ronnie Polidoro and Roshan Sumbaly and Ross Taylor and Ruan Silva and Rui Hou and Rui Wang and Saghar Hosseini and Sahana Chennabasappa and Sanjay Singh and Sean Bell and Seohyun Sonia Kim and Sergey Edunov and Shaoliang Nie and Sharan Narang and Sharath Raparthy and Sheng Shen and Shengye Wan and Shruti Bhosale and Shun Zhang and Simon Vandenhende and Soumya Batra and Spencer Whitman and Sten Sootla and Stephane Collot and Suchin Gururangan and Sydney Borodinsky and Tamar Herman and Tara Fowler and Tarek Sheasha and Thomas Georgiou and Thomas Scialom and Tobias Speckbacher and Todor Mihaylov and Tong Xiao and Ujjwal Karn and Vedanuj Goswami and Vibhor Gupta and Vignesh Ramanathan and Viktor Kerkez and Vincent Gonguet and Virginie Do and Vish Vogeti and Vítor Albiero and Vladan Petrovic and Weiwei Chu and Wenhan Xiong and Wenyin Fu and Whitney Meers and Xavier Martinet and Xiaodong Wang and Xiaofang Wang and Xiaoqing Ellen Tan and Xide Xia and Xinfeng Xie and Xuchao Jia and Xuewei Wang and Yaelle Goldschlag and Yashesh Gaur and Yasmine Babaei and Yi Wen and Yiwen Song and Yuchen Zhang and Yue Li and Yuning Mao and Zacharie Delpierre Coudert and Zheng Yan and Zhengxing Chen and Zoe Papakipos and Aaditya Singh and Aayushi Srivastava and Abha Jain and Adam Kelsey and Adam Shajnfeld and Adithya Gangidi and Adolfo Victoria and Ahuva Goldstand and Ajay Menon and Ajay Sharma and Alex Boesenberg and Alexei Baevski and Allie Feinstein and Amanda Kallet and Amit Sangani and Amos Teo and Anam Yunus and Andrei Lupu and Andres Alvarado and Andrew Caples and Andrew Gu and Andrew Ho and Andrew Poulton and Andrew Ryan and Ankit Ramchandani and Annie Dong and Annie Franco and Anuj Goyal and Aparajita Saraf and Arkabandhu Chowdhury and Ashley Gabriel and Ashwin Bharambe and Assaf Eisenman and Azadeh Yazdan and Beau James and Ben Maurer and Benjamin Leonhardi and Bernie Huang and Beth Loyd and Beto De Paola and Bhargavi Paranjape and Bing Liu and Bo Wu and Boyu Ni and Braden Hancock and Bram Wasti and Brandon Spence and Brani Stojkovic and Brian Gamido and Britt Montalvo and Carl Parker and Carly Burton and Catalina Mejia and Ce Liu and Changhan Wang and Changkyu Kim and Chao Zhou and Chester Hu and Ching-Hsiang Chu and Chris Cai and Chris Tindal and Christoph Feichtenhofer and Cynthia Gao and Damon Civin and Dana Beaty and Daniel Kreymer and Daniel Li and David Adkins and David Xu and Davide Testuggine and Delia David and Devi Parikh and Diana Liskovich and Didem Foss and Dingkang Wang and Duc Le and Dustin Holland and Edward Dowling and Eissa Jamil and Elaine Montgomery and Eleonora Presani and Emily Hahn and Emily Wood and Eric-Tuan Le and Erik Brinkman and Esteban Arcaute and Evan Dunbar and Evan Smothers and Fei Sun and Felix Kreuk and Feng Tian and Filippos Kokkinos and Firat Ozgenel and Francesco Caggioni and Frank Kanayet and Frank Seide and Gabriela Medina Florez and Gabriella Schwarz and Gada Badeer and Georgia Swee and Gil Halpern and Grant Herman and Grigory Sizov and Guangyi and Zhang and Guna Lakshminarayanan and Hakan Inan and Hamid Shojanazeri and Han Zou and Hannah Wang and Hanwen Zha and Haroun Habeeb and Harrison Rudolph and Helen Suk and Henry Aspegren and Hunter Goldman and Hongyuan Zhan and Ibrahim Damlaj and Igor Molybog and Igor Tufanov and Ilias Leontiadis and Irina-Elena Veliche and Itai Gat and Jake Weissman and James Geboski and James Kohli and Janice Lam and Japhet Asher and Jean-Baptiste Gaya and Jeff Marcus and Jeff Tang and Jennifer Chan and Jenny Zhen and Jeremy Reizenstein and Jeremy Teboul and Jessica Zhong and Jian Jin and Jingyi Yang and Joe Cummings and Jon Carvill and Jon Shepard and Jonathan McPhie and Jonathan Torres and Josh Ginsburg and Junjie Wang and Kai Wu and Kam Hou U and Karan Saxena and Kartikay Khandelwal and Katayoun Zand and Kathy Matosich and Kaushik Veeraraghavan and Kelly Michelena and Keqian Li and Kiran Jagadeesh and Kun Huang and Kunal Chawla and Kyle Huang and Lailin Chen and Lakshya Garg and Lavender A and Leandro Silva and Lee Bell and Lei Zhang and Liangpeng Guo and Licheng Yu and Liron Moshkovich and Luca Wehrstedt and Madian Khabsa and Manav Avalani and Manish Bhatt and Martynas Mankus and Matan Hasson and Matthew Lennie and Matthias Reso and Maxim Groshev and Maxim Naumov and Maya Lathi and Meghan Keneally and Miao Liu and Michael L. Seltzer and Michal Valko and Michelle Restrepo and Mihir Patel and Mik Vyatskov and Mikayel Samvelyan and Mike Clark and Mike Macey and Mike Wang and Miquel Jubert Hermoso and Mo Metanat and Mohammad Rastegari and Munish Bansal and Nandhini Santhanam and Natascha Parks and Natasha White and Navyata Bawa and Nayan Singhal and Nick Egebo and Nicolas Usunier and Nikhil Mehta and Nikolay Pavlovich Laptev and Ning Dong and Norman Cheng and Oleg Chernoguz and Olivia Hart and Omkar Salpekar and Ozlem Kalinli and Parkin Kent and Parth Parekh and Paul Saab and Pavan Balaji and Pedro Rittner and Philip Bontrager and Pierre Roux and Piotr Dollar and Polina Zvyagina and Prashant Ratanchandani and Pritish Yuvraj and Qian Liang and Rachad Alao and Rachel Rodriguez and Rafi Ayub and Raghotham Murthy and Raghu Nayani and Rahul Mitra and Rangaprabhu Parthasarathy and Raymond Li and Rebekkah Hogan and Robin Battey and Rocky Wang and Russ Howes and Ruty Rinott and Sachin Mehta and Sachin Siby and Sai Jayesh Bondu and Samyak Datta and Sara Chugh and Sara Hunt and Sargun Dhillon and Sasha Sidorov and Satadru Pan and Saurabh Mahajan and Saurabh Verma and Seiji Yamamoto and Sharadh Ramaswamy and Shaun Lindsay and Shaun Lindsay and Sheng Feng and Shenghao Lin and Shengxin Cindy Zha and Shishir Patil and Shiva Shankar and Shuqiang Zhang and Shuqiang Zhang and Sinong Wang and Sneha Agarwal and Soji Sajuyigbe and Soumith Chintala and Stephanie Max and Stephen Chen and Steve Kehoe and Steve Satterfield and Sudarshan Govindaprasad and Sumit Gupta and Summer Deng and Sungmin Cho and Sunny Virk and Suraj Subramanian and Sy Choudhury and Sydney Goldman and Tal Remez and Tamar Glaser and Tamara Best and Thilo Koehler and Thomas Robinson and Tianhe Li and Tianjun Zhang and Tim Matthews and Timothy Chou and Tzook Shaked and Varun Vontimitta and Victoria Ajayi and Victoria Montanez and Vijai Mohan and Vinay Satish Kumar and Vishal Mangla and Vlad Ionescu and Vlad Poenaru and Vlad Tiberiu Mihailescu and Vladimir Ivanov and Wei Li and Wenchen Wang and Wenwen Jiang and Wes Bouaziz and Will Constable and Xiaocheng Tang and Xiaojian Wu and Xiaolan Wang and Xilun Wu and Xinbo Gao and Yaniv Kleinman and Yanjun Chen and Ye Hu and Ye Jia and Ye Qi and Yenda Li and Yilin Zhang and Ying Zhang and Yossi Adi and Youngjin Nam and Yu and Wang and Yu Zhao and Yuchen Hao and Yundi Qian and Yunlu Li and Yuzi He and Zach Rait and Zachary DeVito and Zef Rosnbrick and Zhaoduo Wen and Zhenyu Yang and Zhiwei Zhao and Zhiyu Ma},
      year={2024},
      eprint={2407.21783},
      archivePrefix={arXiv},
      primaryClass={cs.AI},
      url={https://arxiv.org/abs/2407.21783}, 
}

@article{qwen2024qwen2,
  title={Qwen2.5 Technical Report},
  author={An Yang and Baosong Yang and Beichen Zhang and Binyuan Hui and Bo Zheng and Bowen Yu and Chengyuan Li and Dayiheng Liu and Fei Huang and Guanting Dong and Haoran Wei and Huan Lin and Jian Yang and Jianhong Tu and Jianwei Zhang and Jianxin Yang and Jiaxin Yang and Jingren Zhou and Junyang Lin and Kai Dang and Keming Lu and Keqin Bao and Kexin Yang and Le Yu and Mei Li and Mingfeng Xue and Pei Zhang and Qin Zhu and Rui Men and Runji Lin and Tianhao Li and Tingyu Xia and Xingzhang Ren and Xuancheng Ren and Yang Fan and Yang Su and Yi-Chao Zhang and Yunyang Wan and Yuqi Liu and Zeyu Cui and Zhenru Zhang and Zihan Qiu and Shanghaoran Quan and Zekun Wang},
  journal={ArXiv},
  year={2024},
  volume={abs/2412.15115},
  url={https://api.semanticscholar.org/CorpusID:274859421}
}

@inproceedings{wang-etal-2024-findings,
    title = "Findings of the {WMT} 2024 Shared Task on Discourse-Level Literary Translation",
    author = "Wang, Longyue  and
      Liu, Siyou  and
      Lyu, Chenyang  and
      Jiao, Wenxiang  and
      Wang, Xing  and
      Xu, Jiahao  and
      Tu, Zhaopeng  and
      Gu, Yan  and
      Chen, Weiyu  and
      Wu, Minghao  and
      Zhou, Liting  and
      Koehn, Philipp  and
      Way, Andy  and
      Yuan, Yulin",
    editor = "Haddow, Barry  and
      Kocmi, Tom  and
      Koehn, Philipp  and
      Monz, Christof",
    booktitle = "Proceedings of the Ninth Conference on Machine Translation",
    month = nov,
    year = "2024",
    address = "Miami, Florida, USA",
    publisher = "Association for Computational Linguistics",
    url = "https://aclanthology.org/2024.wmt-1.58/",
    doi = "10.18653/v1/2024.wmt-1.58",
    pages = "699--700"
}

@inproceedings{zhang-etal-2025-good,
    title = "How Good Are {LLM}s for Literary Translation, Really? Literary Translation Evaluation with Humans and {LLM}s",
    author = "Zhang, Ran  and
      Zhao, Wei  and
      Eger, Steffen",
    editor = "Chiruzzo, Luis  and
      Ritter, Alan  and
      Wang, Lu",
    booktitle = "Proceedings of the 2025 Conference of the Nations of the Americas Chapter of the Association for Computational Linguistics: Human Language Technologies (Volume 1: Long Papers)",
    month = apr,
    year = "2025",
    address = "Albuquerque, New Mexico",
    publisher = "Association for Computational Linguistics",
    url = "https://aclanthology.org/2025.naacl-long.548/",
    doi = "10.18653/v1/2025.naacl-long.548",
    pages = "10961--10988",
    ISBN = "979-8-89176-189-6"
}

@inproceedings{neves-etal-2024-findings,
    title = "Findings of the {WMT} 2024 Biomedical Translation Shared Task: Test Sets on Abstract Level",
    author = "Neves, Mariana  and
      Grozea, Cristian  and
      Thomas, Philippe  and
      Roller, Roland  and
      Bawden, Rachel  and
      N{\'e}v{\'e}ol, Aur{\'e}lie  and
      Castle, Steffen  and
      Bonato, Vanessa  and
      Di Nunzio, Giorgio Maria  and
      Vezzani, Federica  and
      Vicente Navarro, Maika  and
      Yeganova, Lana  and
      Jimeno Yepes, Antonio",
    editor = "Haddow, Barry  and
      Kocmi, Tom  and
      Koehn, Philipp  and
      Monz, Christof",
    booktitle = "Proceedings of the Ninth Conference on Machine Translation",
    month = nov,
    year = "2024",
    address = "Miami, Florida, USA",
    publisher = "Association for Computational Linguistics",
    url = "https://aclanthology.org/2024.wmt-1.6/",
    doi = "10.18653/v1/2024.wmt-1.6",
    pages = "124--138"
}

@inproceedings{neves-etal-2023-findings,
    title = "Findings of the {WMT} 2023 Biomedical Translation Shared Task: Evaluation of {C}hat{GPT} 3.5 as a Comparison System",
    author = "Neves, Mariana  and
      Jimeno Yepes, Antonio  and
      N{\'e}v{\'e}ol, Aur{\'e}lie  and
      Bawden, Rachel  and
      Di Nunzio, Giorgio Maria  and
      Roller, Roland  and
      Thomas, Philippe  and
      Vezzani, Federica  and
      Vicente Navarro, Maika  and
      Yeganova, Lana  and
      Wiemann, Dina  and
      Grozea, Cristian",
    editor = "Koehn, Philipp  and
      Haddow, Barry  and
      Kocmi, Tom  and
      Monz, Christof",
    booktitle = "Proceedings of the Eighth Conference on Machine Translation",
    month = dec,
    year = "2023",
    address = "Singapore",
    publisher = "Association for Computational Linguistics",
    url = "https://aclanthology.org/2023.wmt-1.2/",
    doi = "10.18653/v1/2023.wmt-1.2",
    pages = "43--54"
}

@inproceedings{he-etal-2020-box,
    title = "The Box is in the Pen: Evaluating Commonsense Reasoning in Neural Machine Translation",
    author = "He, Jie  and
      Wang, Tao  and
      Xiong, Deyi  and
      Liu, Qun",
    editor = "Cohn, Trevor  and
      He, Yulan  and
      Liu, Yang",
    booktitle = "Findings of the Association for Computational Linguistics: EMNLP 2020",
    month = nov,
    year = "2020",
    address = "Online",
    publisher = "Association for Computational Linguistics",
    url = "https://aclanthology.org/2020.findings-emnlp.327/",
    doi = "10.18653/v1/2020.findings-emnlp.327",
    pages = "3662--3672"
}

@inproceedings{zhang-etal-2023-understanding,
    title = "Understanding and Improving the Robustness of Terminology Constraints in Neural Machine Translation",
    author = "Zhang, Huaao  and
      Wang, Qiang  and
      Qin, Bo  and
      Shi, Zelin  and
      Wang, Haibo  and
      Chen, Ming",
    editor = "Rogers, Anna  and
      Boyd-Graber, Jordan  and
      Okazaki, Naoaki",
    booktitle = "Proceedings of the 61st Annual Meeting of the Association for Computational Linguistics (Volume 1: Long Papers)",
    month = jul,
    year = "2023",
    address = "Toronto, Canada",
    publisher = "Association for Computational Linguistics",
    url = "https://aclanthology.org/2023.acl-long.332/",
    doi = "10.18653/v1/2023.acl-long.332",
    pages = "6029--6042"
}

@article{wang2024retrieval,
  title={Retrieval-Augmented Machine Translation with Unstructured Knowledge},
  author={Wang, Jiaan and Meng, Fandong and Zhang, Yingxue and Zhou, Jie},
  journal={arXiv preprint arXiv:2412.04342},
  year={2024}
}

@inproceedings{yao2024benchmarking,
  title={Benchmarking Machine Translation with Cultural Awareness},
  author={Yao, Binwei and Jiang, Ming and Bobinac, Tara and Yang, Diyi and Hu, Junjie},
  booktitle={Findings of the Association for Computational Linguistics: EMNLP 2024},
  pages={13078--13096},
  year={2024}
}

@inproceedings{rei-etal-2020-comet,
    title = "{COMET}: A Neural Framework for {MT} Evaluation",
    author = "Rei, Ricardo  and
      Stewart, Craig  and
      Farinha, Ana C  and
      Lavie, Alon",
    editor = "Webber, Bonnie  and
      Cohn, Trevor  and
      He, Yulan  and
      Liu, Yang",
    booktitle = "Proceedings of the 2020 Conference on Empirical Methods in Natural Language Processing (EMNLP)",
    month = nov,
    year = "2020",
    address = "Online",
    publisher = "Association for Computational Linguistics",
    url = "https://aclanthology.org/2020.emnlp-main.213/",
    doi = "10.18653/v1/2020.emnlp-main.213",
    pages = "2685--2702"
}

@inproceedings{rei-etal-2022-cometkiwi,
    title = "{C}omet{K}iwi: {IST}-Unbabel 2022 Submission for the Quality Estimation Shared Task",
    author = "Rei, Ricardo  and
      Treviso, Marcos  and
      Guerreiro, Nuno M.  and
      Zerva, Chrysoula  and
      Farinha, Ana C  and
      Maroti, Christine  and
      C. de Souza, Jos{\'e} G.  and
      Glushkova, Taisiya  and
      Alves, Duarte  and
      Coheur, Luisa  and
      Lavie, Alon  and
      Martins, Andr{\'e} F. T.",
    editor = {Koehn, Philipp  and
      Barrault, Lo{\"i}c  and
      Bojar, Ond{\v{r}}ej  and
      Bougares, Fethi  and
      Chatterjee, Rajen  and
      Costa-juss{\`a}, Marta R.  and
      Federmann, Christian  and
      Fishel, Mark  and
      Fraser, Alexander  and
      Freitag, Markus  and
      Graham, Yvette  and
      Grundkiewicz, Roman  and
      Guzman, Paco  and
      Haddow, Barry  and
      Huck, Matthias  and
      Jimeno Yepes, Antonio  and
      Kocmi, Tom  and
      Martins, Andr{\'e}  and
      Morishita, Makoto  and
      Monz, Christof  and
      Nagata, Masaaki  and
      Nakazawa, Toshiaki  and
      Negri, Matteo  and
      N{\'e}v{\'e}ol, Aur{\'e}lie  and
      Neves, Mariana  and
      Popel, Martin  and
      Turchi, Marco  and
      Zampieri, Marcos},
    booktitle = "Proceedings of the Seventh Conference on Machine Translation (WMT)",
    month = dec,
    year = "2022",
    address = "Abu Dhabi, United Arab Emirates (Hybrid)",
    publisher = "Association for Computational Linguistics",
    url = "https://aclanthology.org/2022.wmt-1.60/",
    pages = "634--645"
}

@inproceedings{rei-etal-2022-comet,
    title = "{COMET}-22: Unbabel-{IST} 2022 Submission for the Metrics Shared Task",
    author = "Rei, Ricardo  and
      C. de Souza, Jos{\'e} G.  and
      Alves, Duarte  and
      Zerva, Chrysoula  and
      Farinha, Ana C  and
      Glushkova, Taisiya  and
      Lavie, Alon  and
      Coheur, Luisa  and
      Martins, Andr{\'e} F. T.",
    editor = {Koehn, Philipp  and
      Barrault, Lo{\"i}c  and
      Bojar, Ond{\v{r}}ej  and
      Bougares, Fethi  and
      Chatterjee, Rajen  and
      Costa-juss{\`a}, Marta R.  and
      Federmann, Christian  and
      Fishel, Mark  and
      Fraser, Alexander  and
      Freitag, Markus  and
      Graham, Yvette  and
      Grundkiewicz, Roman  and
      Guzman, Paco  and
      Haddow, Barry  and
      Huck, Matthias  and
      Jimeno Yepes, Antonio  and
      Kocmi, Tom  and
      Martins, Andr{\'e}  and
      Morishita, Makoto  and
      Monz, Christof  and
      Nagata, Masaaki  and
      Nakazawa, Toshiaki  and
      Negri, Matteo  and
      N{\'e}v{\'e}ol, Aur{\'e}lie  and
      Neves, Mariana  and
      Popel, Martin  and
      Turchi, Marco  and
      Zampieri, Marcos},
    booktitle = "Proceedings of the Seventh Conference on Machine Translation (WMT)",
    month = dec,
    year = "2022",
    address = "Abu Dhabi, United Arab Emirates (Hybrid)",
    publisher = "Association for Computational Linguistics",
    url = "https://aclanthology.org/2022.wmt-1.52/",
    pages = "578--585"
}

@inproceedings{kocmi-federmann-2023-large,
    title = "Large Language Models Are State-of-the-Art Evaluators of Translation Quality",
    author = "Kocmi, Tom  and
      Federmann, Christian",
    editor = "Nurminen, Mary  and
      Brenner, Judith  and
      Koponen, Maarit  and
      Latomaa, Sirkku  and
      Mikhailov, Mikhail  and
      Schierl, Frederike  and
      Ranasinghe, Tharindu  and
      Vanmassenhove, Eva  and
      Vidal, Sergi Alvarez  and
      Aranberri, Nora  and
      Nunziatini, Mara  and
      Escart{\'i}n, Carla Parra  and
      Forcada, Mikel  and
      Popovic, Maja  and
      Scarton, Carolina  and
      Moniz, Helena",
    booktitle = "Proceedings of the 24th Annual Conference of the European Association for Machine Translation",
    month = jun,
    year = "2023",
    address = "Tampere, Finland",
    publisher = "European Association for Machine Translation",
    url = "https://aclanthology.org/2023.eamt-1.19/",
    pages = "193--203"
}

@article{lyu2023paradigm,
  title={A paradigm shift: The future of machine translation lies with large language models},
  author={Lyu, Chenyang and Du, Zefeng and Xu, Jitao and Duan, Yitao and Wu, Minghao and Lynn, Teresa and Aji, Alham Fikri and Wong, Derek F and Liu, Siyou and Wang, Longyue},
  journal={arXiv preprint arXiv:2305.01181},
  year={2023}
}

@inproceedings{kocmi-etal-2024-findings,
    title = "Findings of the {WMT}24 General Machine Translation Shared Task: The {LLM} Era Is Here but {MT} Is Not Solved Yet",
    author = "Kocmi, Tom  and
      Avramidis, Eleftherios  and
      Bawden, Rachel  and
      Bojar, Ond{\v{r}}ej  and
      Dvorkovich, Anton  and
      Federmann, Christian  and
      Fishel, Mark  and
      Freitag, Markus  and
      Gowda, Thamme  and
      Grundkiewicz, Roman  and
      Haddow, Barry  and
      Karpinska, Marzena  and
      Koehn, Philipp  and
      Marie, Benjamin  and
      Monz, Christof  and
      Murray, Kenton  and
      Nagata, Masaaki  and
      Popel, Martin  and
      Popovi{\'c}, Maja  and
      Shmatova, Mariya  and
      Steingr{\'i}msson, Steinth{\'o}r  and
      Zouhar, Vil{\'e}m",
    editor = "Haddow, Barry  and
      Kocmi, Tom  and
      Koehn, Philipp  and
      Monz, Christof",
    booktitle = "Proceedings of the Ninth Conference on Machine Translation",
    month = nov,
    year = "2024",
    address = "Miami, Florida, USA",
    publisher = "Association for Computational Linguistics",
    url = "https://aclanthology.org/2024.wmt-1.1/",
    doi = "10.18653/v1/2024.wmt-1.1",
    pages = "1--46"
}

@inproceedings{zhu-etal-2024-multilingual,
    title = "Multilingual Machine Translation with Large Language Models: Empirical Results and Analysis",
    author = "Zhu, Wenhao  and
      Liu, Hongyi  and
      Dong, Qingxiu  and
      Xu, Jingjing  and
      Huang, Shujian  and
      Kong, Lingpeng  and
      Chen, Jiajun  and
      Li, Lei",
    editor = "Duh, Kevin  and
      Gomez, Helena  and
      Bethard, Steven",
    booktitle = "Findings of the Association for Computational Linguistics: NAACL 2024",
    month = jun,
    year = "2024",
    address = "Mexico City, Mexico",
    publisher = "Association for Computational Linguistics",
    url = "https://aclanthology.org/2024.findings-naacl.176/",
    doi = "10.18653/v1/2024.findings-naacl.176",
    pages = "2765--2781"
}

@article{liu2025new,
  title={New trends for modern machine translation with large reasoning models},
  author={Liu, Sinuo and Lyu, Chenyang and Wu, Minghao and Wang, Longyue and Luo, Weihua and Zhang, Kaifu and Shang, Zifu},
  journal={arXiv preprint arXiv:2503.10351},
  year={2025}
}

@article{ye2025well,
  title={How Well Do Large Reasoning Models Translate? A Comprehensive Evaluation for Multi-Domain Machine Translation},
  author={Ye, Yongshi and Fu, Biao and Huang, Chongxuan and Chen, Yidong and Shi, Xiaodong},
  journal={arXiv preprint arXiv:2505.19987},
  year={2025}
}

@inproceedings{cui-etal-2025-multilingual,
    title = "Multilingual Machine Translation with Open Large Language Models at Practical Scale: An Empirical Study",
    author = "Cui, Menglong  and
      Gao, Pengzhi  and
      Liu, Wei  and
      Luan, Jian  and
      Wang, Bin",
    editor = "Chiruzzo, Luis  and
      Ritter, Alan  and
      Wang, Lu",
    booktitle = "Proceedings of the 2025 Conference of the Nations of the Americas Chapter of the Association for Computational Linguistics: Human Language Technologies (Volume 1: Long Papers)",
    month = apr,
    year = "2025",
    address = "Albuquerque, New Mexico",
    publisher = "Association for Computational Linguistics",
    url = "https://aclanthology.org/2025.naacl-long.280/",
    doi = "10.18653/v1/2025.naacl-long.280",
    pages = "5420--5443",
    ISBN = "979-8-89176-189-6"
}

@article{snell2024scaling,
  title={Scaling llm test-time compute optimally can be more effective than scaling model parameters},
  author={Snell, Charlie and Lee, Jaehoon and Xu, Kelvin and Kumar, Aviral},
  journal={arXiv preprint arXiv:2408.03314},
  year={2024}
}

@article{Guo_2025,
   title={DeepSeek-R1 incentivizes reasoning in LLMs through reinforcement learning},
   volume={645},
   ISSN={1476-4687},
   url={http://dx.doi.org/10.1038/s41586-025-09422-z},
   DOI={10.1038/s41586-025-09422-z},
   number={8081},
   journal={Nature},
   publisher={Springer Science and Business Media LLC},
   author={Guo, Daya and Yang, Dejian and Zhang, Haowei and Song, Junxiao and Wang, Peiyi and Zhu, Qihao and Xu, Runxin and Zhang, Ruoyu and Ma, Shirong and Bi, Xiao and Zhang, Xiaokang and Yu, Xingkai and Wu, Yu and Wu, Z. F. and Gou, Zhibin and Shao, Zhihong and Li, Zhuoshu and Gao, Ziyi and Liu, Aixin and Xue, Bing and Wang, Bingxuan and Wu, Bochao and Feng, Bei and Lu, Chengda and Zhao, Chenggang and Deng, Chengqi and Ruan, Chong and Dai, Damai and Chen, Deli and Ji, Dongjie and Li, Erhang and Lin, Fangyun and Dai, Fucong and Luo, Fuli and Hao, Guangbo and Chen, Guanting and Li, Guowei and Zhang, H. and Xu, Hanwei and Ding, Honghui and Gao, Huazuo and Qu, Hui and Li, Hui and Guo, Jianzhong and Li, Jiashi and Chen, Jingchang and Yuan, Jingyang and Tu, Jinhao and Qiu, Junjie and Li, Junlong and Cai, J. L. and Ni, Jiaqi and Liang, Jian and Chen, Jin and Dong, Kai and Hu, Kai and You, Kaichao and Gao, Kaige and Guan, Kang and Huang, Kexin and Yu, Kuai and Wang, Lean and Zhang, Lecong and Zhao, Liang and Wang, Litong and Zhang, Liyue and Xu, Lei and Xia, Leyi and Zhang, Mingchuan and Zhang, Minghua and Tang, Minghui and Zhou, Mingxu and Li, Meng and Wang, Miaojun and Li, Mingming and Tian, Ning and Huang, Panpan and Zhang, Peng and Wang, Qiancheng and Chen, Qinyu and Du, Qiushi and Ge, Ruiqi and Zhang, Ruisong and Pan, Ruizhe and Wang, Runji and Chen, R. J. and Jin, R. L. and Chen, Ruyi and Lu, Shanghao and Zhou, Shangyan and Chen, Shanhuang and Ye, Shengfeng and Wang, Shiyu and Yu, Shuiping and Zhou, Shunfeng and Pan, Shuting and Li, S. S. and Zhou, Shuang and Wu, Shaoqing and Yun, Tao and Pei, Tian and Sun, Tianyu and Wang, T. and Zeng, Wangding and Liu, Wen and Liang, Wenfeng and Gao, Wenjun and Yu, Wenqin and Zhang, Wentao and Xiao, W. L. and An, Wei and Liu, Xiaodong and Wang, Xiaohan and Chen, Xiaokang and Nie, Xiaotao and Cheng, Xin and Liu, Xin and Xie, Xin and Liu, Xingchao and Yang, Xinyu and Li, Xinyuan and Su, Xuecheng and Lin, Xuheng and Li, X. Q. and Jin, Xiangyue and Shen, Xiaojin and Chen, Xiaosha and Sun, Xiaowen and Wang, Xiaoxiang and Song, Xinnan and Zhou, Xinyi and Wang, Xianzu and Shan, Xinxia and Li, Y. K. and Wang, Y. Q. and Wei, Y. X. and Zhang, Yang and Xu, Yanhong and Li, Yao and Zhao, Yao and Sun, Yaofeng and Wang, Yaohui and Yu, Yi and Zhang, Yichao and Shi, Yifan and Xiong, Yiliang and He, Ying and Piao, Yishi and Wang, Yisong and Tan, Yixuan and Ma, Yiyang and Liu, Yiyuan and Guo, Yongqiang and Ou, Yuan and Wang, Yuduan and Gong, Yue and Zou, Yuheng and He, Yujia and Xiong, Yunfan and Luo, Yuxiang and You, Yuxiang and Liu, Yuxuan and Zhou, Yuyang and Zhu, Y. X. and Huang, Yanping and Li, Yaohui and Zheng, Yi and Zhu, Yuchen and Ma, Yunxian and Tang, Ying and Zha, Yukun and Yan, Yuting and Ren, Z. Z. and Ren, Zehui and Sha, Zhangli and Fu, Zhe and Xu, Zhean and Xie, Zhenda and Zhang, Zhengyan and Hao, Zhewen and Ma, Zhicheng and Yan, Zhigang and Wu, Zhiyu and Gu, Zihui and Zhu, Zijia and Liu, Zijun and Li, Zilin and Xie, Ziwei and Song, Ziyang and Pan, Zizheng and Huang, Zhen and Xu, Zhipeng and Zhang, Zhongyu and Zhang, Zhen},
   year={2025},
   month=sep, pages={633–638} }

@misc{jaech2024openai,
      title={OpenAI o1 System Card}, 
      author={Aaron Jaech and Adam Kalai and Adam Lerer and Adam Richardson and Ahmed El-Kishky and Aiden Low and Alec Helyar and Aleksander Madry and Alex Beutel and Alex Carney and Alex Iftimie and Alex Karpenko and Alex Tachard Passos and Alexander Neitz and Alexander Prokofiev and Alexander Wei and Allison Tam and Ally Bennett and Ananya Kumar and Andre Saraiva and Andrea Vallone and Andrew Duberstein and Andrew Kondrich and Andrey Mishchenko and Andy Applebaum and Angela Jiang and Ashvin Nair and Barret Zoph and Behrooz Ghorbani and Ben Rossen and Benjamin Sokolowsky and Boaz Barak and Bob McGrew and Borys Minaiev and Botao Hao and Bowen Baker and Brandon Houghton and Brandon McKinzie and Brydon Eastman and Camillo Lugaresi and Cary Bassin and Cary Hudson and Chak Ming Li and Charles de Bourcy and Chelsea Voss and Chen Shen and Chong Zhang and Chris Koch and Chris Orsinger and Christopher Hesse and Claudia Fischer and Clive Chan and Dan Roberts and Daniel Kappler and Daniel Levy and Daniel Selsam and David Dohan and David Farhi and David Mely and David Robinson and Dimitris Tsipras and Doug Li and Dragos Oprica and Eben Freeman and Eddie Zhang and Edmund Wong and Elizabeth Proehl and Enoch Cheung and Eric Mitchell and Eric Wallace and Erik Ritter and Evan Mays and Fan Wang and Felipe Petroski Such and Filippo Raso and Florencia Leoni and Foivos Tsimpourlas and Francis Song and Fred von Lohmann and Freddie Sulit and Geoff Salmon and Giambattista Parascandolo and Gildas Chabot and Grace Zhao and Greg Brockman and Guillaume Leclerc and Hadi Salman and Haiming Bao and Hao Sheng and Hart Andrin and Hessam Bagherinezhad and Hongyu Ren and Hunter Lightman and Hyung Won Chung and Ian Kivlichan and Ian O'Connell and Ian Osband and Ignasi Clavera Gilaberte and Ilge Akkaya and Ilya Kostrikov and Ilya Sutskever and Irina Kofman and Jakub Pachocki and James Lennon and Jason Wei and Jean Harb and Jerry Twore and Jiacheng Feng and Jiahui Yu and Jiayi Weng and Jie Tang and Jieqi Yu and Joaquin Quiñonero Candela and Joe Palermo and Joel Parish and Johannes Heidecke and John Hallman and John Rizzo and Jonathan Gordon and Jonathan Uesato and Jonathan Ward and Joost Huizinga and Julie Wang and Kai Chen and Kai Xiao and Karan Singhal and Karina Nguyen and Karl Cobbe and Katy Shi and Kayla Wood and Kendra Rimbach and Keren Gu-Lemberg and Kevin Liu and Kevin Lu and Kevin Stone and Kevin Yu and Lama Ahmad and Lauren Yang and Leo Liu and Leon Maksin and Leyton Ho and Liam Fedus and Lilian Weng and Linden Li and Lindsay McCallum and Lindsey Held and Lorenz Kuhn and Lukas Kondraciuk and Lukasz Kaiser and Luke Metz and Madelaine Boyd and Maja Trebacz and Manas Joglekar and Mark Chen and Marko Tintor and Mason Meyer and Matt Jones and Matt Kaufer and Max Schwarzer and Meghan Shah and Mehmet Yatbaz and Melody Y. Guan and Mengyuan Xu and Mengyuan Yan and Mia Glaese and Mianna Chen and Michael Lampe and Michael Malek and Michele Wang and Michelle Fradin and Mike McClay and Mikhail Pavlov and Miles Wang and Mingxuan Wang and Mira Murati and Mo Bavarian and Mostafa Rohaninejad and Nat McAleese and Neil Chowdhury and Neil Chowdhury and Nick Ryder and Nikolas Tezak and Noam Brown and Ofir Nachum and Oleg Boiko and Oleg Murk and Olivia Watkins and Patrick Chao and Paul Ashbourne and Pavel Izmailov and Peter Zhokhov and Rachel Dias and Rahul Arora and Randall Lin and Rapha Gontijo Lopes and Raz Gaon and Reah Miyara and Reimar Leike and Renny Hwang and Rhythm Garg and Robin Brown and Roshan James and Rui Shu and Ryan Cheu and Ryan Greene and Saachi Jain and Sam Altman and Sam Toizer and Sam Toyer and Samuel Miserendino and Sandhini Agarwal and Santiago Hernandez and Sasha Baker and Scott McKinney and Scottie Yan and Shengjia Zhao and Shengli Hu and Shibani Santurkar and Shraman Ray Chaudhuri and Shuyuan Zhang and Siyuan Fu and Spencer Papay and Steph Lin and Suchir Balaji and Suvansh Sanjeev and Szymon Sidor and Tal Broda and Aidan Clark and Tao Wang and Taylor Gordon and Ted Sanders and Tejal Patwardhan and Thibault Sottiaux and Thomas Degry and Thomas Dimson and Tianhao Zheng and Timur Garipov and Tom Stasi and Trapit Bansal and Trevor Creech and Troy Peterson and Tyna Eloundou and Valerie Qi and Vineet Kosaraju and Vinnie Monaco and Vitchyr Pong and Vlad Fomenko and Weiyi Zheng and Wenda Zhou and Wes McCabe and Wojciech Zaremba and Yann Dubois and Yinghai Lu and Yining Chen and Young Cha and Yu Bai and Yuchen He and Yuchen Zhang and Yunyun Wang and Zheng Shao and Zhuohan Li},
      year={2024},
      eprint={2412.16720},
      archivePrefix={arXiv},
      primaryClass={cs.AI},
      url={https://arxiv.org/abs/2412.16720}, 
}

@techreport{openai_o3_o4mini_system_card_2025,
  author       = {{OpenAI}},
  title        = {OpenAI o3 and o4-mini System Card},
  institution  = {OpenAI},
  year         = {2025},
  month        = apr,
  type         = {System Card},
  url          = {https://openai.com/index/o3-o4-mini-system-card/},
  note         = {Published April 16, 2025},
  urldate      = {2025-10-01}
}

@article{muennighoff2025s1,
  title={s1: Simple test-time scaling},
  author={Muennighoff, Niklas and Yang, Zitong and Shi, Weijia and Li, Xiang Lisa and Fei-Fei, Li and Hajishirzi, Hannaneh and Zettlemoyer, Luke and Liang, Percy and Cand{\`e}s, Emmanuel and Hashimoto, Tatsunori},
  journal={arXiv preprint arXiv:2501.19393},
  year={2025}
}

@article{li2025s,
  title={S*: Test time scaling for code generation},
  author={Li, Dacheng and Cao, Shiyi and Cao, Chengkun and Li, Xiuyu and Tan, Shangyin and Keutzer, Kurt and Xing, Jiarong and Gonzalez, Joseph E and Stoica, Ion},
  journal={arXiv preprint arXiv:2502.14382},
  year={2025}
}

@article{wang2025deep,
  title={Deep reasoning translation via reinforcement learning},
  author={Wang, Jiaan and Meng, Fandong and Zhou, Jie},
  journal={arXiv preprint arXiv:2504.10187},
  year={2025}
}

@article{wang2025extrans,
  title={ExTrans: Multilingual Deep Reasoning Translation via Exemplar-Enhanced Reinforcement Learning},
  author={Wang, Jiaan and Meng, Fandong and Zhou, Jie},
  journal={arXiv preprint arXiv:2505.12996},
  year={2025}
}

@misc{he2025r1t1fullyincentivizingtranslation,
      title={R1-T1: Fully Incentivizing Translation Capability in LLMs via Reasoning Learning}, 
      author={Minggui He and Yilun Liu and Shimin Tao and Yuanchang Luo and Hongyong Zeng and Chang Su and Li Zhang and Hongxia Ma and Daimeng Wei and Weibin Meng and Hao Yang and Boxing Chen and Osamu Yoshie},
      year={2025},
      eprint={2502.19735},
      archivePrefix={arXiv},
      primaryClass={cs.CL},
      url={https://arxiv.org/abs/2502.19735}, 
}

@misc{cheng2025seedxbuildingstrongmultilingual,
      title={Seed-X: Building Strong Multilingual Translation LLM with 7B Parameters}, 
      author={Shanbo Cheng and Yu Bao and Qian Cao and Luyang Huang and Liyan Kang and Zhicheng Liu and Yu Lu and Wenhao Zhu and Jingwen Chen and Zhichao Huang and Tao Li and Yifu Li and Huiying Lin and Sitong Liu and Ningxin Peng and Shuaijie She and Lu Xu and Nuo Xu and Sen Yang and Runsheng Yu and Yiming Yu and Liehao Zou and Hang Li and Lu Lu and Yuxuan Wang and Yonghui Wu},
      year={2025},
      eprint={2507.13618},
      archivePrefix={arXiv},
      primaryClass={cs.CL},
      url={https://arxiv.org/abs/2507.13618}, 
}

@article{hurst2024gpt,
  title={GPT-4o System Card},
  author={Aaron Hurst and Adam Lerer and Adam P. Goucher and Adam Perelman and Aditya Ramesh and Aidan Clark and AJ Ostrow and Akila Welihinda and Alan Hayes and Alec Radford and Aleksander Mkadry and Alex Baker-Whitcomb and Alex Beutel and Alex Borzunov and Alex Carney and Alex Chow and Alexander Kirillov and Alex Nichol and Alex Paino and Alex Renzin and Alexandre Passos and Alexander Kirillov and Alexi Christakis and Alexis Conneau and Ali Kamali and Allan Jabri and Allison Moyer and Allison Tam and Amadou Crookes and Amin Tootoochian and Amin Tootoonchian and Ananya Kumar and Andrea Vallone and Andrej Karpathy and Andrew Braunstein and Andrew Cann and Andrew Codispoti and Andrew Galu and Andrew Kondrich and Andrew Tulloch and An-drey Mishchenko and Angela Baek and Angela Jiang and An-toine Pelisse and Antonia Woodford and Anuj Gosalia and Arka Dhar and Ashley Pantuliano and Avi Nayak and Avital Oliver and Barret Zoph and B. Ghorbani and Ben Leimberger and Ben Rossen and Benjamin Sokolowsky and Ben Wang and Benjamin Zweig and Beth Hoover and Blake Samic and Bob McGrew and Bobby Spero and Bogo Giertler and Bowen Cheng and Brad Lightcap and Brandon Walkin and Brendan Quinn and Brian Guarraci and Brian Hsu and Bright Kellogg and Brydon Eastman and Camillo Lugaresi and Carroll L. Wainwright and Cary Bassin and Cary Hudson and Casey Chu and Chad Nelson and Chak Li and Chan Jun Shern and Channing Conger and Charlotte Barette and Chelsea Voss and Chen Ding and Cheng Lu and Chong Zhang and Chris Beaumont and Chris Hallacy and Chris Koch and Christian Gibson and Christina Kim and Christine Choi and Christine McLeavey and Chris Hesse and Claudia Fischer and Clemens Winter and Coley Czarnecki and Colin Jarvis and Colin Wei and Constantin Koumouzelis and Dane Sherburn and Daniel Kappler and Daniel Levin and Daniel Levy and David Carr and David Farhi and David M{\'e}ly and David Robinson and David Sasaki and Denny Jin and Dev Valladares and Dimitris Tsipras and Doug Li and Phong Duc Nguyen and Duncan Findlay and Edede Oiwoh and Edmund Wong and Ehsan Asdar and Elizabeth Proehl and Elizabeth Yang and Eric Antonow and Eric Kramer and Eric Peterson and Eric Sigler and Eric Wallace and Eugene Brevdo and Evan Mays and Farzad Khorasani and Felipe Petroski Such and Filippo Raso and Francis Zhang and Fred von Lohmann and Freddie Sulit and Gabriel Goh and Gene Oden and Geoff Salmon and Giulio Starace and Greg Brockman and Hadi Salman and Hai-Biao Bao and Haitang Hu and Hannah Wong and Haoyu Wang and Heather Schmidt and Heather Whitney and Hee-woo Jun and Hendrik Kirchner and Henrique Pond{\'e} de Oliveira Pinto and Hongyu Ren and Huiwen Chang and Hyung Won Chung and Ian Kivlichan and Ian O’Connell and Ian Osband and Ian Silber and Ian Sohl and Ibrahim Okuyucu and Ikai Lan and Ilya Kostrikov and Ilya Sutskever and Ingmar Kanitscheider and Ishaan Gulrajani and Jacob Coxon and Jacob Menick and Jakub W. Pachocki and James Aung and James Betker and James Crooks and James Lennon and Jamie Ryan Kiros and Jan Leike and Jane Park and Jason Kwon and Jason Phang and Jason Teplitz and Jason Wei and Jason Wolfe and Jay Chen and Jeff Harris and Jenia Varavva and Jessica Gan Lee and Jessica Shieh and Ji Lin and Jiahui Yu and Jiayi Weng and Jie Tang and Jieqi Yu and Joanne Jang and Joaquin Qui{\~n}onero Candela and Joe Beutler and Joe Landers and Joel Parish and Johannes Heidecke and John Schulman and Jonathan Lachman and Jonathan McKay and Jonathan Uesato and Jonathan Ward and Jong Wook Kim and Joost Huizinga and Jordan Sitkin and Jos Kraaijeveld and Joshua Gross and Josh Kaplan and Josh Snyder and Joshua Achiam and Joy Jiao and Joyce Lee and Juntang Zhuang and Justyn Harriman and Kai Fricke and Kai Hayashi and Karan Singhal and Katy Shi and Kavin Karthik and Kayla Wood and Kendra Rimbach and Kenny Hsu and Kenny Nguyen and Keren Gu-Lemberg and Kevin Button and Kevin Liu and Kiel Howe and Krithika Muthukumar and Kyle Luther and Lama Ahmad and Larry Kai and Lauren Itow and Lauren Workman and Leher Pathak and Leo Chen and Li Jing and Lia Guy and Liam Fedus and Liang Zhou and Lien Mamitsuka and Lilian Weng and Lindsay McCallum and Lindsey Held and Ouyang Long and Louis Feuvrier and Lu Zhang and Lukasz Kondraciuk and Lukasz Kaiser and Luke Hewitt and Luke Metz and Lyric Doshi and Mada Aflak and Maddie Simens and Made-laine Boyd and Madeleine Thompson and Marat Dukhan and Mark Chen and Mark Gray and Mark Hudnall and Marvin Zhang and Marwan Aljubeh and Ma-teusz Litwin and Matthew Zeng and Max Johnson and Maya Shetty and Mayank Gupta and Meghan Shah and Mehmet Ali Yatbaz and Mengxue Yang and Mengchao Zhong and Mia Glaese and Mianna Chen and Michael Janner and Michael Lampe and Michael Petrov and Michael Wu and Michele Wang and Michelle Fradin and Michelle Pokrass and Miguel Castro and Miguel Castro and Mikhail Pavlov and Miles Brundage and Miles Wang and Mina Khan and Mira Murati and Mo Bavarian and Molly Lin and Murat Yesildal and Nacho Soto and Natalia Gimelshein and Na-talie Cone and Natalie Staudacher and Natalie Summers and Natan LaFontaine and Neil Chowdhury and Nick Ryder and Nick Stathas and Nick Turley and Nikolas A. Tezak and Niko Felix and Nithanth Kudige and Nitish Shirish Keskar and Noah Deutsch and Noel Bundick and Nora Puckett and Ofir Nachum and Ola Okelola and Oleg Boiko and Oleg Murk and Oliver Jaffe and Olivia Watkins and Olivier Godement and Owen Campbell-Moore and Patrick Chao and Paul McMillan and Pavel Belov and Peng Su and Peter Bak and Peter Bakkum and Peter Deng and Peter Dolan and Peter Hoeschele and Peter Welinder and Phil Tillet and Philip Pronin and Phil Tillet and Prafulla Dhariwal and Qim-ing Yuan and Rachel Dias and Rachel Lim and Rahul Arora and Rajan Troll and Randall Lin and Raphael Gontijo Lopes and Raul Puri and Reah Miyara and Reimar H. Leike and Renaud Gaubert and Reza Zamani and Ricky Wang and Rob Donnelly and Rob Honsby and Rocky Smith and Rohan Sahai and Rohit Ramchandani and Romain Huet and Rory Carmichael and Rowan Zellers and Roy Chen and Ruby Chen and Ruslan Ramilevich Nigmatullin and Ryan Cheu and Saachi Jain and Sam Altman and Sam Schoenholz and Sam Toizer and Samuel Miserendino and Sandhini Agarwal and Sara Culver and Scott Ethersmith and Scott Gray and Sean Grove and Sean Metzger and Shamez Hermani and Shantanu Jain and Shengjia Zhao and Sherwin Wu and Shino Jomoto and Shirong Wu and Shuaiqi Xia and Sonia Phene and Spencer Papay and Srinivas Narayanan and Steve Coffey and Steve Lee and Stewart Hall and Suchir Balaji and Tal Broda and Tal Stramer and Tao Xu and Tarun Gogineni and Taya Christianson and Ted Sanders and Tejal Patwardhan and Thomas Cunninghman and Thomas Degry and Thomas Dimson and Thomas Raoux and Thomas Shadwell and Tianhao Zheng and Todd Underwood and Todor Markov and Toki Sherbakov and Tom Rubin and Tom Stasi and Tomer Kaftan and Tristan Heywood and Troy Peterson and Tyce Walters and Tyna Eloundou and Valerie Qi and Veit Moeller and Vinnie Monaco and Vishal Kuo and Vlad Fomenko and Wayne Chang and Weiyi Zheng and Wenda Zhou and Wesam Manassra and Will Sheu and Wojciech Zaremba and Yash Patil and Yilei Qian and Yongjik Kim and Youlong Cheng and Yu Zhang and Yuchen He and Yuchen Zhang and Yujia Jin and Yunxing Dai and Yury Malkov},
  journal={ArXiv},
  year={2024},
  volume={abs/2410.21276},
  url={https://api.semanticscholar.org/CorpusID:273662196}
}

@article{zheng2025hunyuan,
  title={Hunyuan-MT Technical Report},
  author={Zheng, Mao and Li, Zheng and Qu, Bingxin and Song, Mingyang and Du, Yang and Sun, Mingrui and Wang, Di},
  journal={arXiv preprint arXiv:2509.05209},
  year={2025}
}

@article{rei2025tower+,
  title={Tower+: Bridging Generality and Translation Specialization in Multilingual LLMs},
  author={Rei, Ricardo and Guerreiro, Nuno M and Pombal, Jos{\'e} and Alves, Jo{\~a}o and Teixeirinha, Pedro and Farajian, Amin and Martins, Andr{\'e} FT},
  journal={arXiv preprint arXiv:2506.17080},
  year={2025}
}

@article{tan2025investigating,
  title={Investigating Test-Time Scaling with Reranking for Machine Translation},
  author={Tan, Shaomu and Mitani, Ryosuke and Choudhary, Ritvik and Sekiya, Toshiyuki},
  journal={arXiv preprint arXiv:2509.19020},
  year={2025}
}

@article{son2025linguistic,
  title={Linguistic generalizability of test-time scaling in mathematical reasoning},
  author={Son, Guijin and Hong, Jiwoo and Ko, Hyunwoo and Thorne, James},
  journal={arXiv preprint arXiv:2502.17407},
  year={2025}
}

@article{tran2025scaling,
  title={Scaling Test-time Compute for Low-resource Languages: Multilingual Reasoning in LLMs},
  author={Tran, Khanh-Tung and O'Sullivan, Barry and Nguyen, Hoang D},
  journal={arXiv preprint arXiv:2504.02890},
  year={2025}
}

@article{yong2025crosslingual,
  title={Crosslingual reasoning through test-time scaling},
  author={Yong, Zheng-Xin and Adilazuarda, M Farid and Mansurov, Jonibek and Zhang, Ruochen and Muennighoff, Niklas and Eickhoff, Carsten and Winata, Genta Indra and Kreutzer, Julia and Bach, Stephen H and Aji, Alham Fikri},
  journal={arXiv preprint arXiv:2505.05408},
  year={2025}
}

@inproceedings{feng-etal-2025-tear,
    title = "{TE}a{R}: Improving {LLM}-based Machine Translation with Systematic Self-Refinement",
    author = "Feng, Zhaopeng  and
      Zhang, Yan  and
      Li, Hao  and
      Wu, Bei  and
      Liao, Jiayu  and
      Liu, Wenqiang  and
      Lang, Jun  and
      Feng, Yang  and
      Wu, Jian  and
      Liu, Zuozhu",
    editor = "Chiruzzo, Luis  and
      Ritter, Alan  and
      Wang, Lu",
    booktitle = "Findings of the Association for Computational Linguistics: NAACL 2025",
    month = apr,
    year = "2025",
    address = "Albuquerque, New Mexico",
    publisher = "Association for Computational Linguistics",
    url = "https://aclanthology.org/2025.findings-naacl.218/",
    doi = "10.18653/v1/2025.findings-naacl.218",
    pages = "3922--3938",
    ISBN = "979-8-89176-195-7"
}

@inproceedings{wang-etal-2024-taste,
    title = "{T}as{T}e: Teaching Large Language Models to Translate through Self-Reflection",
    author = "Wang, Yutong  and
      Zeng, Jiali  and
      Liu, Xuebo  and
      Meng, Fandong  and
      Zhou, Jie  and
      Zhang, Min",
    editor = "Ku, Lun-Wei  and
      Martins, Andre  and
      Srikumar, Vivek",
    booktitle = "Proceedings of the 62nd Annual Meeting of the Association for Computational Linguistics (Volume 1: Long Papers)",
    month = aug,
    year = "2024",
    address = "Bangkok, Thailand",
    publisher = "Association for Computational Linguistics",
    url = "https://aclanthology.org/2024.acl-long.333/",
    doi = "10.18653/v1/2024.acl-long.333",
    pages = "6144--6158"
}

@article{liimproving,
  title={Improving Language Model Self-Correction Capability with Meta-Feedback},
  author={Li, Xinnuo and Zhang, Yunxiang and Wang, Lu},
  journal={OpenReview},
  year = "2024",
}

@article{hendy2023good,
  title={How good are gpt models at machine translation? a comprehensive evaluation},
  author={Hendy, Amr and Abdelrehim, Mohamed and Sharaf, Amr and Raunak, Vikas and Gabr, Mohamed and Matsushita, Hitokazu and Kim, Young Jin and Afify, Mohamed and Awadalla, Hany Hassan},
  journal={arXiv preprint arXiv:2302.09210},
  year={2023}
}

@inproceedings{brown2020language,
 author = {Brown, Tom and Mann, Benjamin and Ryder, Nick and Subbiah, Melanie and Kaplan, Jared D and Dhariwal, Prafulla and Neelakantan, Arvind and Shyam, Pranav and Sastry, Girish and Askell, Amanda and Agarwal, Sandhini and Herbert-Voss, Ariel and Krueger, Gretchen and Henighan, Tom and Child, Rewon and Ramesh, Aditya and Ziegler, Daniel and Wu, Jeffrey and Winter, Clemens and Hesse, Chris and Chen, Mark and Sigler, Eric and Litwin, Mateusz and Gray, Scott and Chess, Benjamin and Clark, Jack and Berner, Christopher and McCandlish, Sam and Radford, Alec and Sutskever, Ilya and Amodei, Dario},
 booktitle = {Advances in Neural Information Processing Systems},
 editor = {H. Larochelle and M. Ranzato and R. Hadsell and M.F. Balcan and H. Lin},
 pages = {1877--1901},
 publisher = {Curran Associates, Inc.},
 title = {Language Models are Few-Shot Learners},
 url = {https://proceedings.neurips.cc/paper_files/paper/2020/file/1457c0d6bfcb4967418bfb8ac142f64a-Paper.pdf},
 volume = {33},
 year = {2020}
}

@inproceedings{agrawal-etal-2023-context,
    title = "In-context Examples Selection for Machine Translation",
    author = "Agrawal, Sweta  and
      Zhou, Chunting  and
      Lewis, Mike  and
      Zettlemoyer, Luke  and
      Ghazvininejad, Marjan",
    editor = "Rogers, Anna  and
      Boyd-Graber, Jordan  and
      Okazaki, Naoaki",
    booktitle = "Findings of the Association for Computational Linguistics: ACL 2023",
    month = jul,
    year = "2023",
    address = "Toronto, Canada",
    publisher = "Association for Computational Linguistics",
    url = "https://aclanthology.org/2023.findings-acl.564/",
    doi = "10.18653/v1/2023.findings-acl.564",
    pages = "8857--8873"
}

@inproceedings{vilar-etal-2023-prompting,
    title = "Prompting {P}a{LM} for Translation: Assessing Strategies and Performance",
    author = "Vilar, David  and
      Freitag, Markus  and
      Cherry, Colin  and
      Luo, Jiaming  and
      Ratnakar, Viresh  and
      Foster, George",
    editor = "Rogers, Anna  and
      Boyd-Graber, Jordan  and
      Okazaki, Naoaki",
    booktitle = "Proceedings of the 61st Annual Meeting of the Association for Computational Linguistics (Volume 1: Long Papers)",
    month = jul,
    year = "2023",
    address = "Toronto, Canada",
    publisher = "Association for Computational Linguistics",
    url = "https://aclanthology.org/2023.acl-long.859/",
    doi = "10.18653/v1/2023.acl-long.859",
    pages = "15406--15427"
}

@inproceedings{wu-etal-2025-please,
    title = "Please Translate Again: Two Simple Experiments on Whether Human-Like Reasoning Helps Translation",
    author = "Wu, Di  and
      Aycock, Seth  and
      Monz, Christof",
    editor = "Christodoulopoulos, Christos  and
      Chakraborty, Tanmoy  and
      Rose, Carolyn  and
      Peng, Violet",
    booktitle = "Proceedings of the 2025 Conference on Empirical Methods in Natural Language Processing",
    month = nov,
    year = "2025",
    address = "Suzhou, China",
    publisher = "Association for Computational Linguistics",
    url = "https://aclanthology.org/2025.emnlp-main.1031/",
    doi = "10.18653/v1/2025.emnlp-main.1031",
    pages = "20424--20440",
    ISBN = "979-8-89176-332-6"
}

@inproceedings{kocmi-etal-2025-findings,
    title = "Findings of the {WMT}25 General Machine Translation Shared Task: Time to Stop Evaluating on Easy Test Sets",
    author = "Kocmi, Tom  and
      Artemova, Ekaterina  and
      Avramidis, Eleftherios  and
      Bawden, Rachel  and
      Bojar, Ond{\v{r}}ej  and
      Dranch, Konstantin  and
      Dvorkovich, Anton  and
      Dukanov, Sergey  and
      Fishel, Mark  and
      Freitag, Markus  and
      Gowda, Thamme  and
      Grundkiewicz, Roman  and
      Haddow, Barry  and
      Karpinska, Marzena  and
      Koehn, Philipp  and
      Lakougna, Howard  and
      Lundin, Jessica  and
      Monz, Christof  and
      Murray, Kenton  and
      Nagata, Masaaki  and
      Perrella, Stefano  and
      Proietti, Lorenzo  and
      Popel, Martin  and
      Popovi{\'c}, Maja  and
      Riley, Parker  and
      Shmatova, Mariya  and
      Steingr{\'i}msson, Steinth{\'o}r  and
      Yankovskaya, Lisa  and
      Zouhar, Vil{\'e}m",
    editor = "Haddow, Barry  and
      Kocmi, Tom  and
      Koehn, Philipp  and
      Monz, Christof",
    booktitle = "Proceedings of the Tenth Conference on Machine Translation",
    month = nov,
    year = "2025",
    address = "Suzhou, China",
    publisher = "Association for Computational Linguistics",
    url = "https://aclanthology.org/2025.wmt-1.22/",
    doi = "10.18653/v1/2025.wmt-1.22",
    pages = "355--413",
    ISBN = "979-8-89176-341-8"
}

\appendix

\section{Evaluation Prompts}
\label{sec:prompts}

\noindent\textbf{GRB Prompt:}
\begin{promptbox}
Score the following translation from \{src\_lang\} to \{tgt\_lang\} with respect to the human reference on a continuous scale from 0 to 100, where a score of zero means ``no meaning preserved'' and score of one hundred means ``perfect preservation of meaning, with faithfulness, expressiveness, and elegance''.\\
Only output the score number.\\

\{src\_lang\} source: ``\{src\_text\}''\\
\{tgt\_lang\} human reference: ``\{tgt\_text\}''\\
\{tgt\_lang\} translation: ``\{hyp\_text\}''\\

Score:
\end{promptbox}

\noindent\textbf{GRF Prompt:}
\begin{promptbox}
Score the following translation from \{src\_lang\} to \{tgt\_lang\} on a continuous scale from 0 to 100, where a score of zero means ``no meaning preserved'' and score of one hundred means ``perfect preservation of meaning, with faithfulness, expressiveness, and elegance''.\\
Only output the score number.\\

\{src\_lang\} source: ``\{src\_text\}''\\
\{tgt\_lang\} translation: ``\{hyp\_text\}''\\

Score:
\end{promptbox}

\noindent\textbf{GEA\(_{100}\) Prompt:}
\begin{promptbox}
Please evaluate the following \{tgt\_lang\} translation of an \{src\_lang\} text. 
Rate the translation on a scale of 0 to 100, where: \\

- 10 points: Poor translation; the text is somewhat understandable but contains significant errors and awkward phrasing that greatly hinder comprehension for a \{tgt\_lang\} reader. \\
- 30 points: Fair translation; the text conveys the basic meaning but lacks fluency and contains several awkward phrases or inaccuracies, making it challenging for a \{tgt\_lang\} reader to fully grasp the intended message. \\
- 50 points: Good translation; the text is mostly fluent and conveys the original meaning well, but may have minor awkwardness or slight inaccuracies that could confuse a \{tgt\_lang\} reader. \\
- 70 points: Very good translation; the text is smooth and natural, effectively conveying the intended meaning, but may still have minor issues that could slightly affect understanding for a \{tgt\_lang\} reader. \\
- 90 points: Excellent translation; the text is fluent and natural, conveying the original meaning clearly and effectively, with no significant issues that would hinder understanding for a \{tgt\_lang\} reader. \\

Please only output the score number.
\end{promptbox}

\noindent\textbf{GEA\(_{5}\) Prompt:}
\begin{promptbox}
Please evaluate the following \{tgt\_lang\} translation of an \{src\_lang\} text. 
Rate the translation on a scale of 0 to 5, where: \\

- 1 point: Poor translation; the text is somewhat understandable but contains significant errors and awkward phrasing that greatly hinder comprehension for a \{tgt\_lang\} reader. \\
- 2 points: Fair translation; the text conveys the basic meaning but lacks fluency and contains several awkward phrases or inaccuracies, making it challenging for a \{tgt\_lang\} reader to fully grasp the intended message. \\
- 3 points: Good translation; the text is mostly fluent and conveys the original meaning well, but may have minor awkwardness or slight inaccuracies that could confuse a \{tgt\_lang\} reader. \\
- 4 points: Very good translation; the text is smooth and natural, effectively conveying the intended meaning, but may still have minor issues that could slightly affect understanding for a \{tgt\_lang\} reader. \\
- 5 points: Excellent translation; the text is fluent and natural, conveying the original meaning clearly and effectively, with no significant issues that would hinder understanding for a \{tgt\_lang\} reader. \\

Please only output the score number.
\end{promptbox}

\section{Evaluation Results of General-purpose Models}
Tables~\ref{tab:avg_comet}–\ref{tab:avg_grf} summarize the average performance of the General-purpose models across all datasets with respect to the COMET, COMETKiwi, GRB, and GRF metrics, respectively.

\begin{table}[ht]
\centering
\resizebox{\columnwidth}{!}{%
\begin{tabular}{lccccccccc}
\toprule
\multirow{2}{*}{Model} & \multicolumn{7}{c}{Budget} \\
\cmidrule(lr){2-10}
 & {0} & {100} & {200} & {300} & {500} & {1000} & {2000} & {Low} & {High} \\
\midrule
Cogito-3B & 0.712 & 0.671 & 0.695 & 0.698 & 0.704 & 0.707 & 0.704 &  &  \\
Cogito-8B & 0.765 & 0.762 & 0.764 & 0.763 & 0.762 & 0.768 & 0.769 &  &  \\
Qwen3-0.6B & 0.702 & 0.694 & 0.699 & 0.695 & 0.695 & 0.696 & 0.689 &  &  \\
Qwen3-1.7B & 0.760 & 0.757 & 0.764 & 0.763 & 0.764 & 0.769 & 0.764 &  &  \\
Qwen3-4B & 0.789 & 0.785 & 0.789 & 0.790 & 0.791 & 0.791 & 0.787 &  &  \\
Qwen3-8B & 0.801 & 0.800 & 0.799 & 0.800 & 0.799 & 0.798 & 0.797 &  &  \\
Qwen3-14B & 0.805 & 0.807 & 0.803 & 0.806 & 0.805 & 0.799 & 0.797 &  &  \\
Qwen3-32B & 0.804 & 0.804 & 0.802 & 0.803 & 0.802 & 0.802 & 0.799 &  &  \\
Grok-3-Mini &  &  &  &  &  &  &  & 0.794 & 0.795 \\
\bottomrule
\end{tabular}%
}
\caption{Average COMET scores of general-purpose models across all datasets with varying thinking budgets.}
\label{tab:avg_comet}
\end{table}

\begin{table}[ht]
\centering
\resizebox{\columnwidth}{!}{%
\begin{tabular}{lccccccccc}
\toprule
\multirow{2}{*}{Model} & \multicolumn{7}{c}{Budget} \\
\cmidrule(lr){2-10}
 & {0} & {100} & {200} & {300} & {500} & {1000} & {2000} & {Low} & {High} \\
\midrule
Cogito-3B & 0.634 & 0.599 & 0.619 & 0.625 & 0.629 & 0.631 & 0.632 &  &  \\
Cogito-8B & 0.677 & 0.676 & 0.681 & 0.679 & 0.680 & 0.682 & 0.682 &  &  \\
Qwen3-0.6B & 0.618 & 0.617 & 0.616 & 0.617 & 0.614 & 0.615 & 0.609 &  &  \\
Qwen3-1.7B & 0.675 & 0.671 & 0.678 & 0.678 & 0.679 & 0.685 & 0.676 &  &  \\
Qwen3-4B & 0.706 & 0.702 & 0.705 & 0.707 & 0.709 & 0.709 & 0.700 &  &  \\
Qwen3-8B & 0.719 & 0.718 & 0.717 & 0.718 & 0.717 & 0.718 & 0.712 &  &  \\
Qwen3-14B & 0.725 & 0.724 & 0.723 & 0.725 & 0.725 & 0.718 & 0.713 &  &  \\
Qwen3-32B & 0.724 & 0.720 & 0.721 & 0.722 & 0.724 & 0.723 & 0.717 &  &  \\
Grok-3-Mini &  &  &  &  &  &  &  & 0.701 & 0.701 \\
\bottomrule
\end{tabular}%
}
\caption{Average COMETKiwi scores of general-purpose models across all datasets with varying thinking budgets.}
\label{tab:avg_cometkiwi}
\end{table}

\begin{table}[ht]
\centering
\resizebox{\columnwidth}{!}{%
\begin{tabular}{lccccccccc}
\toprule
\multirow{2}{*}{Model} & \multicolumn{7}{c}{Budget} \\
\cmidrule(lr){2-10}
 & {0} & {100} & {200} & {300} & {500} & {1000} & {2000} & {low} & {high} \\
\midrule
Cogito-3B & 82.546 & 78.016 & 79.590 & 80.571 & 81.176 & 81.497 & 81.937 &  &  \\
Cogito-8B & 88.177 & 88.552 & 88.312 & 88.402 & 87.948 & 88.432 & 88.351 &  &  \\
Qwen3-0.6B & 58.165 & 57.507 & 57.518 & 57.813 & 57.317 & 57.372 & 57.734 &  &  \\
Qwen3-1.7B & 74.422 & 74.634 & 75.522 & 74.883 & 75.076 & 75.678 & 75.753 &  &  \\
Qwen3-4B & 84.909 & 84.921 & 85.361 & 85.809 & 86.165 & 86.379 & 86.148 &  &  \\
Qwen3-8B & 89.204 & 89.394 & 89.441 & 89.623 & 89.651 & 89.541 & 90.001 &  &  \\
Qwen3-14B & 90.899 & 90.875 & 90.707 & 91.075 & 91.163 & 91.253 & 91.294 &  &  \\
Qwen3-32B & 90.949 & 91.103 & 91.204 & 91.333 & 91.340 & 91.556 & 91.455 &  &  \\
Grok-3-Mini &  &  &  &  &  &  &  & 92.529 & 92.451 \\
\bottomrule
\end{tabular}%
}
\caption{Average GRB scores of general-purpose models across all datasets with varying thinking budgets.}
\label{tab:avg_grb}
\end{table}

\begin{table}[ht]
\centering
\resizebox{\columnwidth}{!}{%
\begin{tabular}{lccccccccc}
\toprule
\multirow{2}{*}{Model} & \multicolumn{7}{c}{Budget} \\
\cmidrule(lr){2-10}
 & {0} & {100} & {200} & {300} & {500} & {1000} & {2000} & {low} & {high} \\
\midrule
Cogito-3B & 80.928 & 77.468 & 80.178 & 81.680 & 81.577 & 81.843 & 80.632 &  &  \\
Cogito-8B & 89.470 & 89.550 & 88.707 & 88.942 & 88.082 & 89.067 & 89.284 &  &  \\

Qwen3-0.6B & 56.858 & 56.943 & 57.549 & 57.137 & 57.232 & 56.890 & 56.990 &  &  \\
Qwen3-1.7B & 74.954 & 75.377 & 76.210 & 75.829 & 75.723 & 76.770 & 76.591 &  &  \\
Qwen3-4B & 86.238 & 86.229 & 86.722 & 86.965 & 87.603 & 87.806 & 87.577 &  &  \\
Qwen3-8B & 90.321 & 90.625 & 90.648 & 90.816 & 91.024 & 91.123 & 91.258 &  &  \\
Qwen3-14B & 92.084 & 92.016 & 91.932 & 92.359 & 92.503 & 92.492 & 92.547 &  &  \\
Qwen3-32B & 92.122 & 92.268 & 92.404 & 92.608 & 92.702 & 92.835 & 92.782 &  &  \\
Grok-3-Mini &  &  &  &  &  &  &  & 93.494 & 93.594 \\
\bottomrule
\end{tabular}%
}
\caption{Average GRF scores of general-purpose models across all datasets with varying thinking budgets.}
\label{tab:avg_grf}
\end{table}

\section{Evaluation Results and Thinking Token Statistics of DRT Models on Literary Translation Tasks}
\begin{figure}[ht]
    \centering
    \includegraphics[width=\columnwidth]{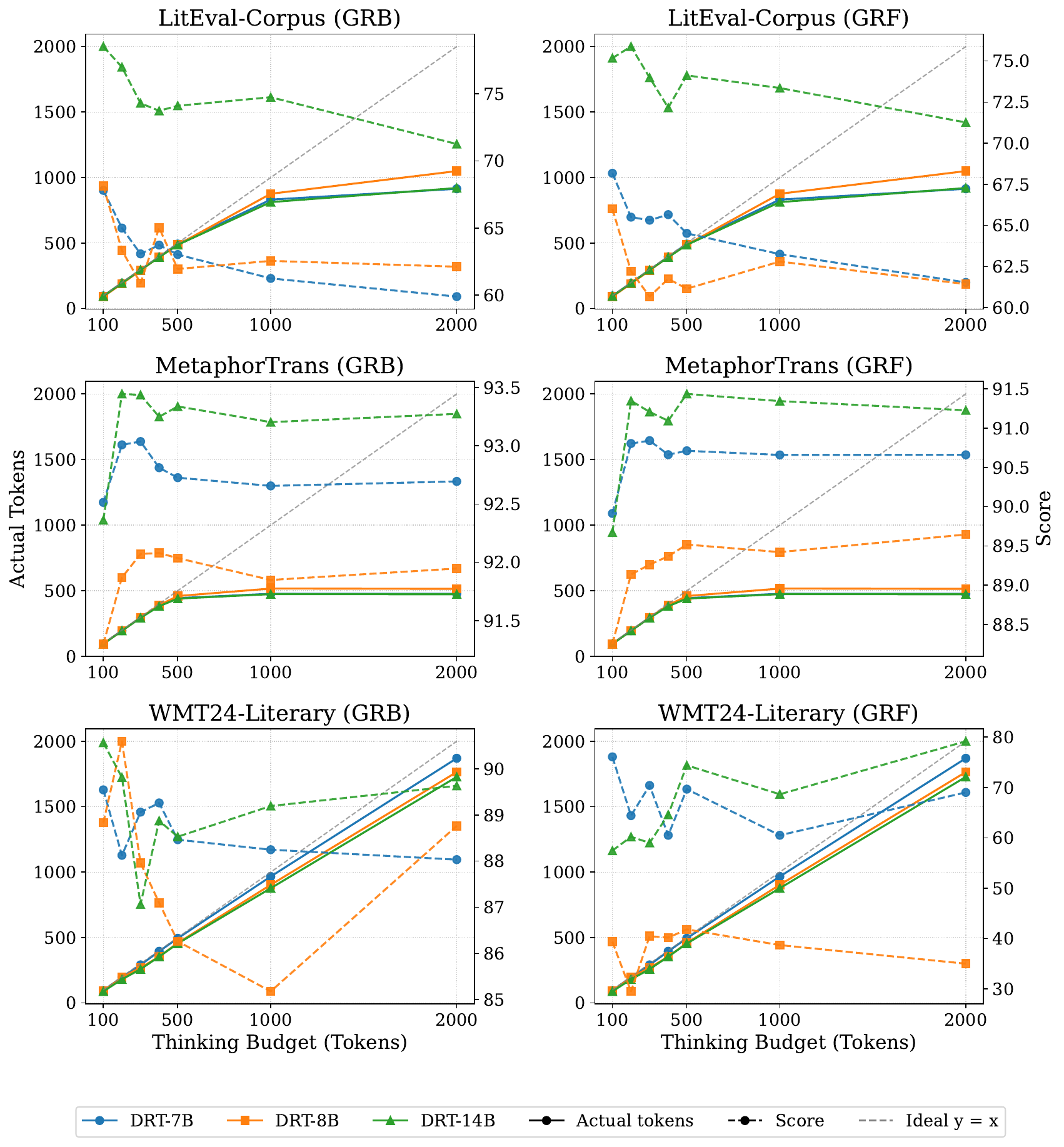}
    \caption{Performance and actual generated thinking tokens of DRT models across 3 literary translation tasks. 
    }
    \label{fig:drt_budget_vs_actual_grb_grf_combined}
\end{figure}

Tables~\ref{tab:drt_gea_100}-~\ref{tab:drt_grf} present the performance of the DRT models across three literary translation benchmarks with respect to the GEA$_{100}$, GEA$_{5}$, GRB, and GRF metrics, respectively.
Table~\ref {tab:drt_actual_tokens} shows the token statistics under different budgets.
Figure~\ref{fig:drt_budget_vs_actual_grb_grf_combined} visualizes DRT models' translation quality (GRB\&GRF score, right y-axis) and the actual number of thinking tokens (left y-axis). On the in-domain MetaphorTrans task, there is a positive growth of the quality scores as the thinking budget increases to around 300 tokens, then the score generally stabilizes. While on the other two out-of-domain tasks, the performance fluctuates and shows an overall downward trend as thinking tokens increase. 

\begin{table}[ht]
\centering
\resizebox{\columnwidth}{!}{%
\begin{tabular}{l l c c c c c c c}
\toprule
\multirow{2}{*}{Task} & \multirow{2}{*}{Model} & \multicolumn{7}{c}{Budget} \\
\cmidrule(lr){3-9}
 &  & 100 & 200 & 300 & 400 & 500 & 1000 & 2000 \\
\midrule
\multirow{3}{*}{MetaphorTrans} 
  & DRT-7B & 75.08 & 79.63 & 80.52 & 80.72 & 80.81 & 80.74 & 80.89 \\
  & DRT-8B  & 71.81 & 75.36 & 76.57 & 77.48 & 77.70 & 77.89 & 78.38 \\
  & DRT-14B & 76.88 & 81.02 & 81.62 & 81.90 & 82.16 & 82.39 & 82.45 \\
\midrule
\multirow{3}{*}{LitEval-Corpus}
  & DRT-7B & 65.09 & 63.66 & 65.07 & 64.38 & 62.75 & 68.08 & 64.47 \\
  & DRT-8B  & 58.11 & 56.92 & 55.00 & 55.39 & 56.10 & 57.98 & 58.26 \\
  & DRT-14B & 67.94 & 69.90 & 68.97 & 73.83 & 74.65 & 76.07 & 73.57 \\
\midrule
\multirow{3}{*}{WMT24-Literary} 
  & DRT-7B & 79.45 & 75.48 & 74.50 & 77.35 & 78.03 & 77.15 & 76.62 \\
  & DRT-8B  & 63.00 & 59.66 & 61.57 & 59.75 & 67.05 & 61.18 & 57.16 \\
  & DRT-14B & 75.34 & 74.38 & 77.18 & 74.06 & 75.54 & 81.37 & 80.01 \\
\bottomrule
\end{tabular}}
\caption{GEA$_{100}$ scores of DRT models on literary translation tasks with varying thinking budgets.}
\label{tab:drt_gea_100}
\end{table}

\begin{table}[ht]
\centering
\resizebox{\columnwidth}{!}{%
\begin{tabular}{l l c c c c c c c}
\toprule
\multirow{2}{*}{Task} & \multirow{2}{*}{Model} & \multicolumn{7}{c}{Budget} \\
\cmidrule(lr){3-9}
 &  & 100 & 200 & 300 & 400 & 500 & 1000 & 2000 \\
\midrule
\multirow{3}{*}{MetaphorTrans} 
  & DRT-7B & 3.75 & 3.92 & 3.95 & 3.96 & 3.96 & 3.96 & 3.95 \\
  & DRT-8B  & 3.58 & 3.75 & 3.81 & 3.85 & 3.86 & 3.87 & 3.88 \\
  & DRT-14B & 3.83 & 3.97 & 3.98 & 4.00 & 4.00 & 4.00 & 4.01 \\
\midrule
\multirow{3}{*}{LitEval-Corpus} 
  & DRT-7B & 3.27 & 3.26 & 3.27 & 3.31 & 3.12 & 3.29 & 3.23 \\
  & DRT-8B  & 3.02 & 2.95 & 2.85 & 2.96 & 2.96 & 3.08 & 3.01 \\
  & DRT-14B & 3.42 & 3.45 & 3.37 & 3.43 & 3.58 & 3.55 & 3.56 \\
\midrule
\multirow{3}{*}{WMT24-Literary} 
  & DRT-7B & 4.02 & 3.85 & 3.65 & 3.58 & 3.83 & 3.83 & 3.64 \\
  & DRT-8B  & 2.67 & 2.93 & 2.90 & 3.13 & 3.09 & 3.01 & 2.60 \\
  & DRT-14B & 3.73 & 3.73 & 3.58 & 3.79 & 3.57 & 3.86 & 3.94 \\
\bottomrule
\end{tabular}}
\caption{GEA$_{5}$ scores of DRT models on literary translation tasks with varying thinking budgets.}
\label{tab:drt_gea_5}
\end{table}

\begin{table}[ht]
\centering
\resizebox{\columnwidth}{!}{%
\begin{tabular}{l l c c c c c c c}
\toprule
\multirow{2}{*}{Task} & \multirow{2}{*}{Model} & \multicolumn{7}{c}{Budget} \\
\cmidrule(lr){3-9}
 &  & 100 & 200 & 300 & 400 & 500 & 1000 & 2000 \\
\midrule
\multirow{3}{*}{MetaphorTrans} 
  & DRT-7B & 92.51 & 93.01 & 93.04 & 92.81 & 92.73 & 92.66 & 92.69 \\
  & DRT-8B  & 91.30 & 91.87 & 92.07 & 92.08 & 92.04 & 91.85 & 91.95 \\
  & DRT-14B & 92.36 & 93.45 & 93.44 & 93.25 & 93.34 & 93.20 & 93.27 \\
\midrule
\multirow{3}{*}{LitEval-Corpus} 
  & DRT-7B & 67.81 & 65.00 & 63.08 & 63.73 & 63.02 & 61.25 & 59.89 \\
  & DRT-8B  & 68.13 & 63.33 & 60.89 & 65.02 & 61.95 & 62.55 & 62.12 \\
  & DRT-14B & 78.52 & 76.99 & 74.30 & 73.73 & 74.11 & 74.75 & 71.26 \\
\midrule
\multirow{3}{*}{WMT24-Literary} 
  & DRT-7B & 89.54 & 88.13 & 89.06 & 89.26 & 88.46 & 88.25 & 88.03 \\
  & DRT-8B  & 88.83 & 90.59 & 87.96 & 87.10 & 86.26 & 85.18 & 88.76 \\
  & DRT-14B & 90.56 & 89.81 & 87.06 & 88.87 & 88.53 & 89.19 & 89.63 \\
\bottomrule
\end{tabular}}
\caption{GRB scores of DRT models on literary translation tasks with varying thinking budgets.}
\label{tab:drt_grb}
\end{table}

\begin{table}[ht]
\centering
\resizebox{\columnwidth}{!}{%
\begin{tabular}{l l c c c c c c c}
\toprule
\multirow{2}{*}{Task} & \multirow{2}{*}{Model} & \multicolumn{7}{c}{Budget} \\
\cmidrule(lr){3-9}
 &  & 100 & 200 & 300 & 400 & 500 & 1000 & 2000 \\
\midrule
\multirow{3}{*}{MetaphorTrans} 
  & DRT-7B & 89.91 & 90.81 & 90.84 & 90.66 & 90.71 & 90.66 & 90.66 \\
  & DRT-8B  & 88.25 & 89.14 & 89.26 & 89.37 & 89.52 & 89.42 & 89.65 \\
  & DRT-14B & 89.67 & 91.35 & 91.21 & 91.09 & 91.44 & 91.34 & 91.23 \\
\midrule
\multirow{3}{*}{LitEval-Corpus} 
  & DRT-7B & 68.17 & 65.50 & 65.32 & 65.66 & 64.52 & 63.25 & 61.53 \\
  & DRT-8B  & 66.01 & 62.22 & 60.67 & 61.75 & 61.15 & 62.80 & 61.43 \\
  & DRT-14B & 75.18 & 75.88 & 73.99 & 72.16 & 74.13 & 73.36 & 71.27 \\
\midrule
\multirow{3}{*}{WMT24-Literary} 
  & DRT-7B & 76.06 & 64.40 & 70.39 & 60.50 & 69.66 & 60.51 & 69.00 \\
  & DRT-8B  & 39.38 & 29.54 & 40.48 & 40.19 & 41.83 & 38.72 & 35.03 \\
  & DRT-14B & 57.46 & 60.20 & 59.01 & 64.58 & 74.38 & 68.64 & 79.16 \\
\bottomrule
\end{tabular}}
\caption{GRF scores of DRT models on literary translation tasks with varying thinking budgets.}
\label{tab:drt_grf}
\end{table}

\begin{table}[ht]
\centering
\resizebox{\columnwidth}{!}{%
\begin{tabular}{l l c c c c c c c}
\toprule
\multirow{2}{*}{Task} & \multirow{2}{*}{Model} & \multicolumn{7}{c}{Budget} \\
\cmidrule(lr){3-9}
 &  & 100 & 200 & 300 & 400 & 500 & 1000 & 2000 \\
\midrule
\multirow{3}{*}{MetaphorTrans} 
  & DRT-7B & 96.50 & 195.74 & 295.37 & 387.48 & 443.83 & 476.93 & 476.94 \\
  & DRT-8B  & 95.50 & 194.87 & 294.83 & 390.00 & 459.57 & 516.67 & 514.38 \\
  & DRT-14B & 94.27 & 195.28 & 293.20 & 382.22 & 439.87 & 474.03 & 472.84 \\
\midrule
\multirow{3}{*}{LitEval-Corpus} 
  & DRT-7B & 95.97 & 193.87 & 293.37 & 395.45 & 492.04 & 831.38 & 913.88 \\
  & DRT-8B  & 91.12 & 189.58 & 290.99 & 393.68 & 486.73 & 876.08 & 1049.21 \\
  & DRT-14B & 94.90 & 194.57 & 294.72 & 391.62 & 486.15 & 812.47 & 919.86 \\
\midrule
\multirow{3}{*}{WMT24-Literary} 
  & DRT-7B & 95.89 & 195.91 & 290.76 & 394.55 & 493.93 & 966.17 & 1870.02 \\
  & DRT-8B  & 90.71 & 196.56 & 268.99 & 353.15 & 458.29 & 902.59 & 1764.82 \\
  & DRT-14B & 88.43 & 180.07 & 258.03 & 353.63 & 453.92 & 876.82 & 1727.59 \\
\bottomrule
\end{tabular}}
\caption{Actual thinking tokens of DRT models on literary translation tasks with varying thinking budgets.}
\label{tab:drt_actual_tokens}
\end{table}

\section{Post-editing Prompts and Detailed Results}
\label{sec:post-editing-prompts}

\subsection{Prompts}
We experiment with two variants of post-editing prompts: with and without an additional quality score (QS). Examples are provided below.

\noindent\textbf{Post-editing with QS:}
\begin{promptbox}
You are a professional translator, and your task is to refine the \{tgt\_lang\} draft translation below based on the \{src\_lang\} source text and its quality evaluation.

Please only provide me with the refined translation, without any additional explanations.

Source Text: \{src\_text\}

Draft Translation: \{hyp\_text\}

Quality Score: \{quality\_score\}/100
\end{promptbox}

\noindent\textbf{Post-editing without QS:}
\begin{promptbox}
You are a professional translator, and your task is to refine the \{tgt\_lang\} draft translation below based on the \{src\_lang\} source text.

Please only provide me with the refined translation, without any additional explanations.

Source Text: \{src\_text\}

Draft Translation: \{hyp\_text\}
\end{promptbox}

\subsection{Detailed Results}
Tables~\ref{tab:post-editing-grb} and \ref{tab:post-editing-grf} report the GRB and GRF scores respectively. 

\begin{table*}[ht]
\centering
\resizebox{\textwidth}{!}{%
\begin{tabular}{l c c c c c c c c}
\toprule
\multirow{2}{*}{Model} & \multirow{2}{*}{Original} &
\multicolumn{3}{c}{No QS} & \multicolumn{3}{c}{QS} \\
\cmidrule(lr){3-5} \cmidrule(lr){6-8}
& & {Budget=0} & {Budget=500} & {Budget=1000} & {Budget=0} & {Budget=500} & {Budget=1000} \\
\midrule
Qwen3-0.6B & 69.293 & 69.756\,(+0.463) & 69.417\,(+0.124) & 69.445\,(+0.152) & 69.112\,(-0.181) & 69.538\,(+0.246) & 69.410\,(+0.117) \\
Qwen3-1.7B & 82.327 & 80.913\,(-1.415) & 82.896\,(+0.568) & 82.930\,(+0.602) & 81.438\,(-0.889) & 82.878\,(+0.551) & 83.275\,(+0.947) \\
Qwen3-4B  & 88.695 & 88.623\,(-0.072) & 89.734\,(+1.039) & 89.888\,(+1.193) & 88.883\,(+0.188) & 89.725\,(+1.030) & 89.835\,(+1.140) \\
Qwen3-8B  & 91.253 & 91.275\,(+0.022) & 91.618\,(+0.365) & 91.747\,(+0.494) & 91.173\,(-0.080) & 91.653\,(+0.400) & 91.657\,(+0.404) \\
Qwen3-14B & 92.199 & 92.241\,(+0.042) & 92.482\,(+0.283) & 92.529\,(+0.330) & 92.294\,(+0.095) & 92.448\,(+0.249) & 92.472\,(+0.272) \\
Qwen3-32B & 92.197 & 92.415\,(+0.219) & 92.502\,(+0.305) & 92.497\,(+0.301) & 92.365\,(+0.169) & 92.282\,(+0.085) & 92.334\,(+0.138) \\
\bottomrule
\end{tabular}%
}
\caption{Post-editing GRB score with and without QS at different budgets.}
\label{tab:post-editing-grb}
\end{table*}

\begin{table*}[ht]
\centering
\resizebox{\textwidth}{!}{%
\begin{tabular}{l c c c c c c c c}
\toprule
\multirow{2}{*}{Model} & \multirow{2}{*}{Original} &
\multicolumn{3}{c}{No QS} & \multicolumn{3}{c}{QS} \\
\cmidrule(lr){3-5} \cmidrule(lr){6-8}
& & {Budget=0} & {Budget=500} & {Budget=1000} & {Budget=0} & {Budget=500} & {Budget=1000} \\
\midrule
Qwen3-0.6B & 67.477 & 67.757\,(+0.281) & 67.936\,(+0.459) & 67.989\,(+0.512) & 67.987\,(+0.510) & 68.085\,(+0.608) & 68.057\,(+0.580) \\
Qwen3-1.7B & 81.727 & 79.159\,(-2.568) & 81.988\,(+0.261) & 82.305\,(+0.578) & 80.260\,(-1.467) & 82.445\,(+0.718) & 82.730\,(+1.003) \\
Qwen3-4B  & 88.588 & 88.625\,(+0.038) & 89.936\,(+1.349) & 90.057\,(+1.470) & 88.755\,(+0.168) & 89.885\,(+1.298) & 90.031\,(+1.444) \\
Qwen3-8B  & 91.208 & 90.440\,(-0.768) & 92.007\,(+0.799) & 92.018\,(+0.810) & 89.499\,(-1.709) & 92.051\,(+0.843) & 91.933\,(+0.725) \\
Qwen3-14B & 92.261 & 92.613\,(+0.352) & 92.763\,(+0.501) & 92.853\,(+0.592) & 92.642\,(+0.381) & 92.757\,(+0.495) & 92.810\,(+0.549) \\
Qwen3-32B & 92.256 & 92.767\,(+0.511) & 92.923\,(+0.667) & 92.864\,(+0.608) & 92.816\,(+0.560) & 92.768\,(+0.512) & 92.899\,(+0.643) \\
\bottomrule
\end{tabular}%
}
\caption{Post-editing GRF score with and without QS at different budgets.}
\label{tab:post-editing-grf}
\end{table*}

\end{document}